\documentclass[journal]{IEEEtran}

\usepackage{siunitx}								
\usepackage{epsfig} 									
\usepackage{mathptmx} 								
\usepackage{times} 									
\usepackage{amsmath} 								
\usepackage{amssymb,amsfonts} 					
\usepackage{flushend} 								
\usepackage{cite}
\usepackage{placeins}
\usepackage{graphicx}
\usepackage{color,soul}
\usepackage{graphicx}
\usepackage{algorithm2e}
\usepackage{color}
\usepackage{verbatim}
\usepackage{array}
\usepackage{wrapfig}
\usepackage{flushend}
\usepackage{subfigure}
\usepackage{paralist}
\usepackage{mathtools}
\usepackage{multirow}

\newcommand{\specialcell}[2][c]{%
 \begin{tabular}[#1]{@{}c@{}}#2\end{tabular}}
\begin{document}
\title{Beyond RMSE: Do machine-learned models of road user interaction produce human-like behavior?}
\author{Aravinda Ramakrishnan Srinivasan$^{1*}$, Yi-Shin Lin$^{1}$, Morris Antonello$^{2}$, Anthony Knittel$^{2}$, Mohamed Hasan$^{3}$, Majd Hawasly$^{2}$, John Redford$^{2}$, Subramanian Ramamoorthy$^{2,4}$, Matteo Leonetti$^{5}$, Jac Billington$^{6}$, \\Richard Romano$^{1}$, and Gustav Markkula$^{1*}$%
\thanks{This project has received funding from UK Engineering and Physical Sciences Research Council under fellowship named COMMOTIONS - \\Computational Models of Traffic Interactions for Testing of Automated Vehicles - EP/S005056/1. For the purpose of open access, the author(s) has applied a Creative Commons Attribution (CC BY) license to any Accepted Manuscript version arising.}%
\thanks{$^{1}$ Institute for Transport Studies, University of Leeds, UK}%
\thanks{$^{2}$ Five AI Ltd., UK}%
\thanks{$^{3}$ School of Computing, University of Leeds, UK}%
\thanks{$^{4}$ Institute of Perception, Action and Behaviour, School of Informatics, \\University of Edinburgh, UK}%
\thanks{$^{5}$ Department of Informatics, King's College London, UK}%
\thanks{$^{6}$ School of Psychology, University of Leeds, UK}%
\thanks{$^{*}$ Corresponding authors: {\tt\small A.R.Srinivasan@leeds.ac.uk, G.Markkula@leeds.ac.uk}}%
}

\markboth{IEEE Transactions on Intelligent Transportation Systems}{Srinivasan \MakeLowercase{\textit{et al.}}: Beyond RMSE: Do machine-learned models of road user interaction produce human-like behavior?}
\maketitle

\begin{abstract}
Autonomous vehicles use a variety of sensors and machine-learned models to predict the behavior of surrounding road users. Most of the machine-learned models in the literature focus on quantitative error metrics like the root mean square error (RMSE) to learn and report their models' capabilities. This focus on quantitative error metrics tends to ignore the more important behavioral aspect of the models, raising the question of whether these models really predict human-like behavior. Thus, we propose to analyze the output of machine-learned models much like we would analyze human data in conventional behavioral research. We introduce quantitative metrics to demonstrate presence of three different behavioral phenomena in a naturalistic highway driving dataset: 1) The kinematics-dependence of who passes a merging point first 2) Lane change by an on-highway vehicle to accommodate an on-ramp vehicle 3) Lane changes by vehicles on the highway to avoid lead vehicle conflicts. Then, we analyze the behavior of three machine-learned models using the same metrics. Even though the models' RMSE value differed, all the models captured the kinematic-dependent merging behavior but struggled at varying degrees to capture the more nuanced courtesy lane change and highway lane change behavior. Additionally, the collision aversion analysis during lane changes showed that the models struggled to capture the physical aspect of human driving: leaving adequate gap between the vehicles. Thus, our analysis highlighted the inadequacy of simple quantitative metrics and the need to take a broader behavioral perspective when analyzing machine-learned models of human driving predictions.
\end{abstract}

\section{Introduction}
Roadways are used by traffic actors with varying capabilities: motor vehicles and vulnerable road users such as pedestrians and cyclists. The motor vehicles can be further categorized depending on their capabilities according to the Society of Automotive Engineers (SAE)~\cite{SAEloa}. Vehicles of different levels of autonomy are going to coexist in the roadways before fully automated driving (SAE Level 5) takes over every vehicle on the roadways. This presents an interesting and challenging task for vehicles with increased driving automation: they have to coexist with humans both as drivers and other vulnerable road users. Thus, the autonomous driving system needs to learn human road user behaviors to be an effective and safe participant in their interactions with humans. The need for understanding human road user behaviors is also evident from some of the early accidents~\cite{davies2018google,stilgoe2020killed,claybrook2018autonomous} caused by vehicles with increased autonomous capabilities. Managing to interact safely and appropriately with human road users also has the added benefit of improving the public's perception of autonomous driving technologies if they are more human-like/human-interpretive~\cite{hilgarter2020public,madigan2016acceptance}.

Generally, autonomous vehicles use machine-learning algorithms' to detect and predict other road users trajectories. A quantitative metric like the root mean square error (RMSE) is used to report the accuracy of machine-learning algorithms' prediction~\cite{Alahi_2016_CVPR,deo2018convolutional,8917228,Ma_Zhu_Zhang_Yang_Wang_Manocha_2019}. The RMSE measures the average displacement error between the predicted trajectories and their corresponding ground truth. This can be either the average error over the entire prediction horizon, reported as mADE, mean average displacement error in literature~\cite{Sun_2022_CVPR,Phan-Minh_2020_CVPR,Ettinger_2021_ICCV,ngiam2021scene} or average of momentary error at different time points in the prediction horizon, reported as mFDE, mean final displacement error in literature~\cite{Sun_2022_CVPR,Phan-Minh_2020_CVPR,Ettinger_2021_ICCV,ngiam2021scene}. A model with the lowest error is generally considered the better alternative. In cases where the algorithm predicts a probability distribution over a set of maneuvers with spatial trajectory uncertainty, the preferred quantitative metric is the negative log likelihood (NLL) along with RMSE values. The NLL is a measure of how close the predicted probability distribution is to the ground truth, the lower the NLL value the closer is the fit to ground truth. Another approach in this context is to use classification-based metrics such as mean average precision (mAP) and miss rate (MR)~\cite{Ettinger_2021_ICCV,Sun_2022_CVPR,ngiam2021scene}.

The strength of all these high-level quantitative metrics is that they allow straightforward model comparison. However, by reducing the complexity of human road user behavior to a single number, they necessarily sacrifice detail and lose the qualitative aspect of behavior, leaving several questions open, such as: How low an average trajectory prediction error is low enough? What if a model is getting good prediction scores on these metrics by optimizing some aspects of behavior which are not very important to human acceptance and safety while completely missing some aspects that are? Below, we provide an overview of relevant literature in this area, before defining the specific objectives of this paper.

\subsection{Literature review}
In a typical roadway, different road users have to interact with each other. These interactions have been formally defined as situations where the behaviors of at least two of the involved road users can be interpreted as being influenced by a space-sharing conflict between them~\cite{markkula2020interaction}. In order to facilitate safe interaction between autonomous vehicles and human road users, there have been attempts to model road user behaviors from different perspectives, ranging from modeling pedestrian crossing decisions~\cite{camara2020pedestrian,camara2020pedestrian2} to attempting derivation of a driver model from recorded human driven trajectories~\cite{claussmann2019review}.

One way to predict road user behaviors is to formulate trajectory prediction as time-series model. Recurrent neural networks (RNNs) have been used to model time series data. Long Short-Term Memory (LSTM) is an RNN model that has been used to predict pedestrian trajectories while taking into account the neighbor agents' trajectories~\cite{Alahi_2016_CVPR}. Deo and Trivedi improved this idea by adding a convolution based social pooling (CSP) layer in order to preserve the spatial relationship between vehicles for predicting human driven vehicles trajectories~\cite{deo2018convolutional}. This CSP-LSTM network architecture has since become a benchmark for newer models~\cite{Ma_Zhu_Zhang_Yang_Wang_Manocha_2019,8917228,GSANYe2022}, where the RMSE is used to quantify and compare the prediction accuracy. Other architectures based on graph neural networks (GNN)~\cite{GSANYe2022,lee2019joint} proved to be competitive and further reduced state-of-the-art trajectory errors metric like RMSE, improving over convolution-based architectures.

Another way to model the trajectory prediction task is to consider it as a classification problem. One of the best performing models with this approach utilized a map-based approach to generate candidate trajectories along with prelabeled interacting agents from the dataset as input to the classifier. The classifier outputs the trajectory prediction with highest classifier confidence~\cite{Phan-Minh_2020_CVPR}. Like previous trajectories prediction models, they also showcased the performance with high-level average quantitative metrics like mADE, mFDE. A comprehensive literature review of deep-learning based trajectory prediction algorithms is presented by Mozaffari et al.~\cite{mozaffari2020deep}.

Recently, there have been works focused on behavioral analysis of the machine-learned prediction models. Zgonnikov et al.~\cite{zgonnikov2022modeling} conducted an interdisciplinary workshop focused on how human robot interaction (HRI) can be improved by utilizing human behavior modeling. The importance of understanding the behaviors of the artificial intelligence (AI) including machine learning (ML) algorithms since they are becoming ubiquitous was argued in~\cite{rahwan2019machine}. Herman et al.~\cite{herman2021pedestrian} concentrated on the requirements of pedestrian behavior prediction with autonomous driving as the underlying application. Siebinga et al.~\cite{siebinga2021validating} highlighted the need to validate the underlying driver models in the interaction-aware controllers (IACs). They proposed a two stage behavioral validation of the underlying human driver behavior model on which the autonomous driving algorithm rely to be aware of, and plan for, human driver intentions and behaviors. The first stage was tactical behavior analysis which included maneuvers like lane changes, car following; and the second stage was operational safety analysis: whether the maneuvers were executed in a safe manner. Karle et al.~\cite{9733973} present an extensive review of scenario understanding and motion prediction.

\subsection{Problem definition}

Despite the recent interest in a more behavioral understanding of ML models, an important question remains open: Are average trajectory error metrics (like RMSE, mFDE, mADE, NLL, etc.) a good way of comparing different models' behavioral competence? To answer that question, we need to both measure an average trajectory error metric and do the behavioral analysis for multiple machine-learned models. Thus, for a comprehensive behavioral analysis of existing and new machine-learned models, first behaviors need to be quantitatively defined and observed in naturalistic driving. Then, all machine-learned models can be compared fairly by examining whether the same behaviors can be observed in their predictions. This, in addition to the average trajectory error metrics comparison, can give a clear picture regarding the models' behavioral capabilities.

However, taking a more behavioral perspective forces one to engage with the high complexity of human road user behavior which otherwise gets hidden behind the high-level metrics. A very large number of human road user interaction phenomena have been identified in empirical research~\cite{siebinga2021validating, markkula2020interaction,markkula_lin_srinivasan_billington_leonetti_kalantari_yang_lee_madigan_merat_2022}. In this work, we have constrained our analyses to highway driving, to keep a feasible scope. Highway driving has naturally occurring interesting interactions where two or more road users end up affecting each other's decision, such as the merging scenarios where vehicles on the highway and on-ramp interact in order to facilitate a safe passage for everyone involved. Additionally, highway driving scenario was preferred because of ready availability of existing data and models.

The highway merging scenario has been the subject of research from a white box (non machine-learned) perspective extensively~\cite{choudhury_modeling_2009,kang2017game}. These models suggest that when a vehicle involved has a clear kinematic lead over another one, the vehicle with kinematic lead will pass the merging point first. In situations where the kinematic lead is ambiguous (Figure~\ref{fig:KLA_illu}), it is dependent on other factors like social convention, competition, or cooperation between the drivers involved to determine who will pass the merging location first. Similar behavior has been observed in interaction between pedestrian(s)~\cite{olivier2013collision}.

Another interesting behavior that has been studied and modeled in detail is the lane change behavior of the highway vehicle in order to accommodate the on-ramp vehicle (Figure~\ref{fig:MLC_illu})~\cite{zheng2014recent}. The lane change by the highway vehicle in a merging situation can be attributed to different goals like the highway vehicle preference to keep its speed~\cite{kondyli2009driver}, the merging vehicles' aggressive driving style leading the highway vehicle to switch lane for safety, or the merging vehicle waiting for a suitable gap to emerge and merge into the traffic safely~\cite{choudhury_modeling_2009}. In this paper, following the precedence set in our previous work~\cite{srinivasan2021comparing}, we are interested in identifying and analyzing the ``courtesy'' lane changes by the highway vehicle(s) to facilitate safe merging for the on-ramp vehicle(s).

Finally, the lane change behavior of vehicles not in the outermost lane of the highway into the highway to maintain their kinematic edge or to accelerate (Figure~\ref{fig:HLC_illu}) is another interesting behavior which has been of interest and modeled by researchers~\cite{7728060,5701973,jula2000,Nilsson2016}. To the best of our knowledge, none of these three phenomena (tendency of kinematically leading vehicle to pass the nominal merging point first during merging situations, tendency of on-highway vehicle(s) to make ``courtesy'' lane change in order to accommodate on-ramp vehicle(s), and tendency of on-highway vehicles to switch lanes into the highway to preserve/improve their kinematic edge) have been explicitly analyzed either in naturalistic trajectory data or in machine-learned trajectory prediction models.

In summary, the primary goal for this work is to explore a new quantitative method that takes into account the qualitative behavioral aspects of human driving behavior for comparison of different machine-learned models. This should augment the existing traditional quantitative metrics, like RMSE, to provide a better comparison metric to fairly analyze the predictions of machine-learned models. In order to achieve this, we mathematically define the above-mentioned three behaviors and other details of our analysis method. First, we observe the behaviors in the naturalistic driving data. Next, we train the machine-learned models with the same dataset and observe the behavioral patterns exhibited with the same analysis method. This allows for a fair appraisal of the machine-learned models' behavioral capabilities. We have previously presented an early version of this work in a conference paper~\cite{srinivasan2021comparing} where we performed qualitative tests of a single machine-learned model for two behavioral phenomena. In this paper, we introduce tests for one additional behavioral phenomenon, extend our qualitative analyses with quantitative metrics as well as a behaviorally informed safety analysis. Moreover, we apply our behavioral analysis on two additional machine-learned models, allowing us to also demonstrate how our method can be used for model comparison.

\section{EXPERIMENTAL METHODS}
\subsection{Dataset}
The driving trajectories from the Next Generation SIMulation (NGSIM) dataset~\cite{NGSiMData} include both highway driving and urban driving. We used only the highway driving trajectories. The highway driving portion consists of data collected at two different highway locations in USA, the US101 and I80 highways. The same dataset has been used for training the CSP-LSTM machine-learned model in~\cite{deo2018convolutional}. The two new machine-learned models (details in Section~\ref{sec:IPBT} and Section~\ref{sec:BIP-KM}) are previously unpublished and in order to facilitate a fair comparison between CSP-LSTM and them, we trained all three models with the same dataset and with the 70-10-20 (training-validation-test) split recommended in the publicly available CSP-LSTM code~\cite{deo2018convolutional}.

There was a total of 1,268 unique vehicles on the I80 and 1,667 unique vehicles on the US101 in the test and validation set. Out of these, $147$ and $111$ vehicles used the on-ramp to enter the highway, respectively. A total of $419$ and $653$ lane changes were made from outermost highway lane into the highway (during merging scenario) in US101 and I80 highways, respectively. An aggregate of $1180$ and $1635$ lane change into nominal faster lane by highway driving vehicles (excluding the ones already accounted for in merging scenario) were made in the respective highways. For the CSP-LSTM model, we verified that our trained network achieved similar RMSE performance to that reported in~\cite{deo2018convolutional}. In this paper, all the presented behavioral comparisons were made between the naturalistic trajectories and the machine-learned models output for only the validation and the test splits of the dataset.

\subsection{Machine-learned models}
In this section we provide a brief summary of the three machine-learned models. The space constraint prevented us from providing full details about the machine-learned models and thus should not be considered as comprehensive definition of them. Since the focus of this paper is to investigate the possible benefits of more behaviorally oriented model comparison rather than advance the state of the art in trajectory prediction, the specific choice of machine-learned models is less important here; they merely serve as three examples of different approaches to the trajectory prediction task.

\subsubsection{Convolutional Social Pooling - Long Short Term Memory neural network (CSP-LSTM)}
\begin{figure}
 \centering
 \includegraphics[width=1.0\linewidth, height = 5cm]{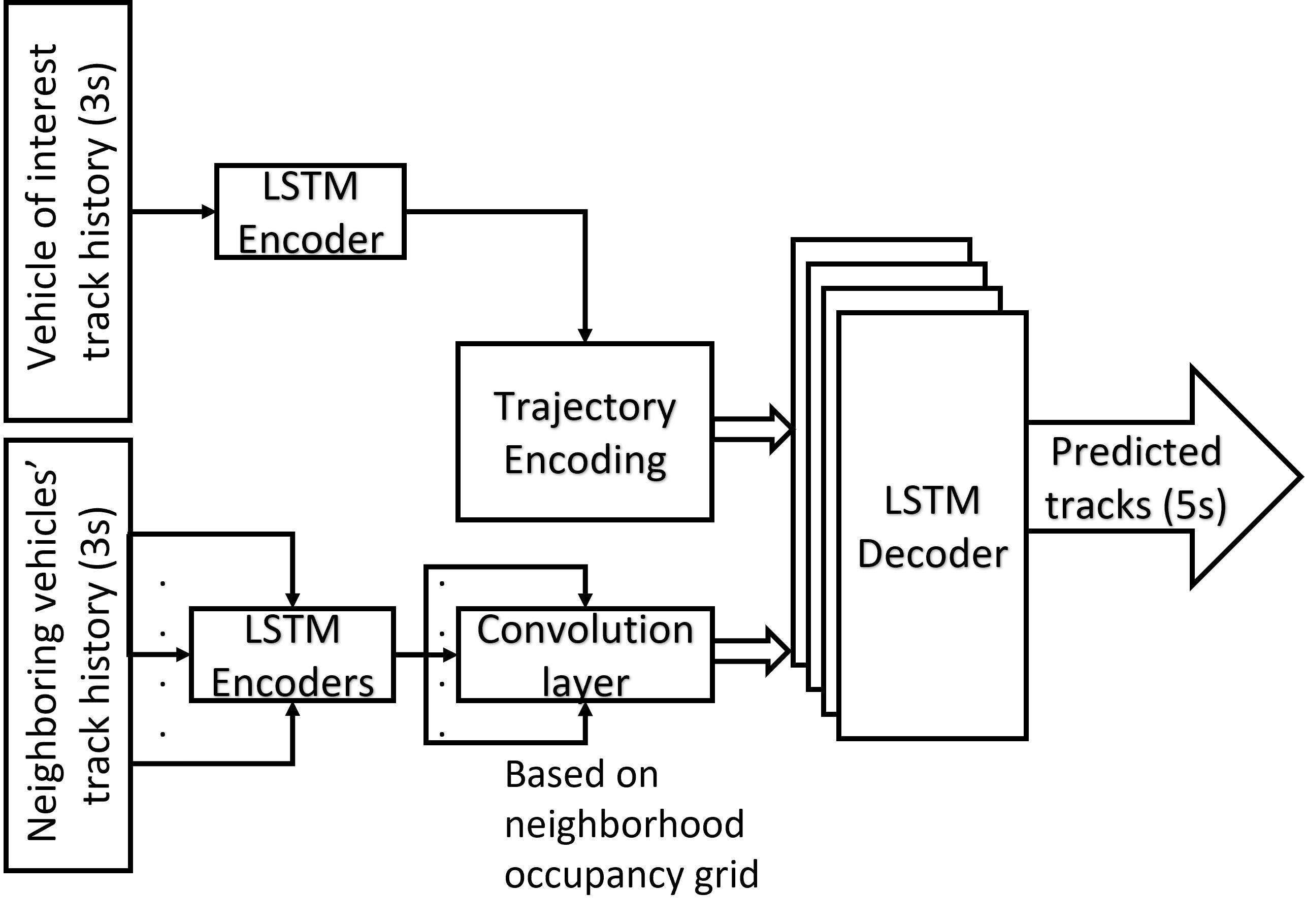}
 \caption{CSP-LSTM network architecture block diagram}
 \label{fig:CSP-LSTM}
 \vspace{-0.5cm}
\end{figure}

The CSP-LSTM model architecture for vehicle trajectory prediction task was introduced by Deo and Trivedi~\cite{deo2018convolutional}. A flow chart of the different components involved is presented in \figurename~\ref{fig:CSP-LSTM}. The CSP-LSTM model is fed a fixed $3$ seconds at $2.5$~$Hz$ trajectory input of the vehicle of interest (ego vehicle), along with the trajectories of the neighboring vehicles within a $13 \times 3$ grid where the ego vehicle is centered in the grid. The trajectories of neighboring vehicles are encoded and the spatial relation between them is preserved through the utilization of the convolution layer. It is trained on the human-driven trajectories and it outputs $5$ seconds prediction (at $2.5$~$Hz$). The error between the prediction and the actual trajectories are back propagated to train the model.

\subsubsection{Interactive Prediction from Basis Trajectories (IPBT)} \label{sec:IPBT}
\begin{figure}
 \centering
 \includegraphics[width=1.0\linewidth]{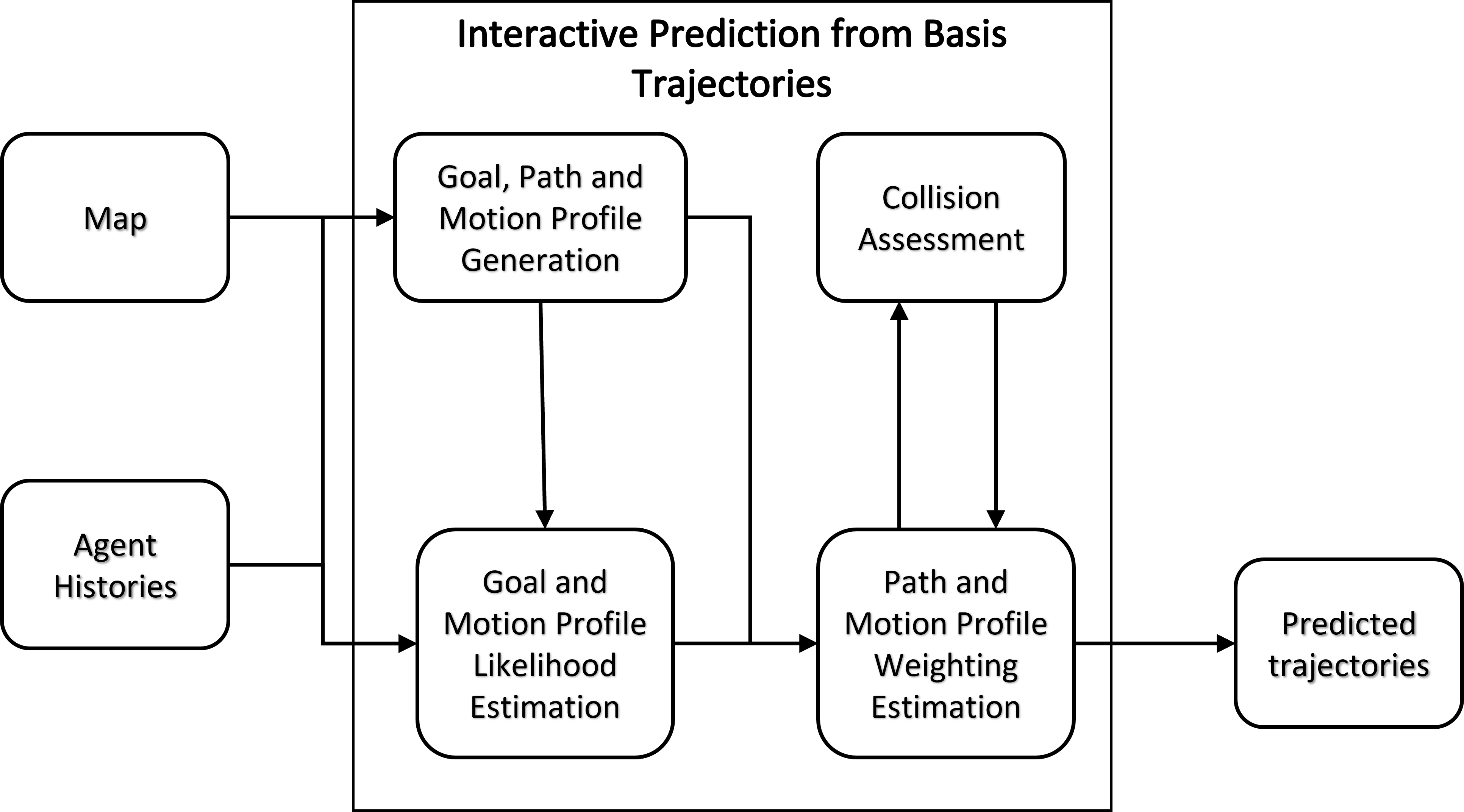}
 \caption{Interactive Prediction from Basis Trajectories (IPBT) block diagram}
 \label{fig:InteractivePredictionOverview}
 \vspace{-0.5cm}
\end{figure}

The IPBT method consists of a series of processing steps: goal and motion profile likelihood estimation, generation of candidate paths and motion profiles, estimation of candidate trajectories, collision estimation between agent proposals, and revised trajectory estimation. An overview of the IPBT model is shown in \figurename~\ref{fig:InteractivePredictionOverview}.

First, goal likelihood estimation is performed using inverse planning~\cite{baker2009action},~\cite{knittel2022dipa}, by comparing the historic trajectory of a vehicle against a number of proposed paths towards goals from a historic time step. Additionally, motion profile likelihood estimation is performed by generating a number of proposed motion profiles from a historic vehicle kinematic state, and comparing the historic motion against the proposed motion profiles. Goal and motion profile likelihood estimation are used to influence the estimates of future paths and motion profiles, based on shared parameters like common goal or speed target. This is implemented based on a multi-layer perceptron (MLP) network.

Next, candidate paths are generated from the map by fitting splines between the start position and the goal via way-points. The map is a topological representation of the driveable region, a set of evenly spaced way-points along lane mid-lines. Candidate motion profiles are produced by identifying target kinematic states from the road layout, the surrounding agents and from a number of behavior parameters, and fitting speed profiles between the initial and target states using splines. Trajectories are generated from the basis set of candidate paths and motion profiles using a weighting function implemented as a convolutional neural network. This uses the goal and motion profile likelihood estimation, historic agent states, and future candidate path and profiles to produce predicted trajectories as a weighted combination of the candidate paths and motion profiles. The model also outputs spatial uncertainty for each time step and prediction mode, represented with an elliptical (2D) Gaussian.

Finally, to produce predictions based on interactions between agents, collisions between the predicted modes of the various agents are assessed and used to produce a refined prediction. Refinements are introduced through multiple levels of prediction, which receive inputs of initial predictions and produce subsequent agent predictions that consider the interactions between the predicted actions of the various agents. In this way the interactive prediction model produces estimates of trajectories, spatial uncertainty and mode probability for each agent that take into account the expected behaviors of other agents, to produce predictions that are influenced by interactions.

\subsubsection{Bayesian Inverse Planning with Kinematic Models (BIP-KM)} \label{sec:BIP-KM}
This model performs multi-modal prediction by taking a Bayesian inverse planning approach~\cite{baker2009action}. Similar to IPBT, in order to predict the future states of an agent, the model begins by hypothesizing a map features-based collection of goals. It then calculates a physically feasible plan for how the agent might reach any given goal and it assesses the likelihoods by comparing the plans to the agent's past observations from perception. Finally, a joint distribution over goals and trajectories is inferred by inverting the planning model using Bayesian inference, integrating the likelihood with the prior. In contrast to deep neural network-based predictors such as~\cite{mersch2021maneuver, tang2019multiple, deo2018convolutional} and IPBT, this approach forms an interpretable algorithm. Relying on maps, physics-based models and trajectory generation algorithms, it generalizes to new environments and situations without additional training data while satisfying physical realism guaranteed by construction. The model parameters have a physical meaning and they can be tuned on data sets without sacrificing interpretability. In contrast to other neural network based methods such as CSP-LSTM or~\cite{mersch2021maneuver, tang2019multiple}, and in contrast to other inverse planning implementations such as~\cite{albrecht2021interpretable,antonello2022flash} or IPBT, this implementation predicts each agent independently for reduced computational requirements and simplicity. Traffic context, i.e. the motion and relative spatial configuration of neighboring agents, or future interactions with other agents are not considered.

\begin{figure}
 \centering
 \includegraphics[width=1.0\linewidth]{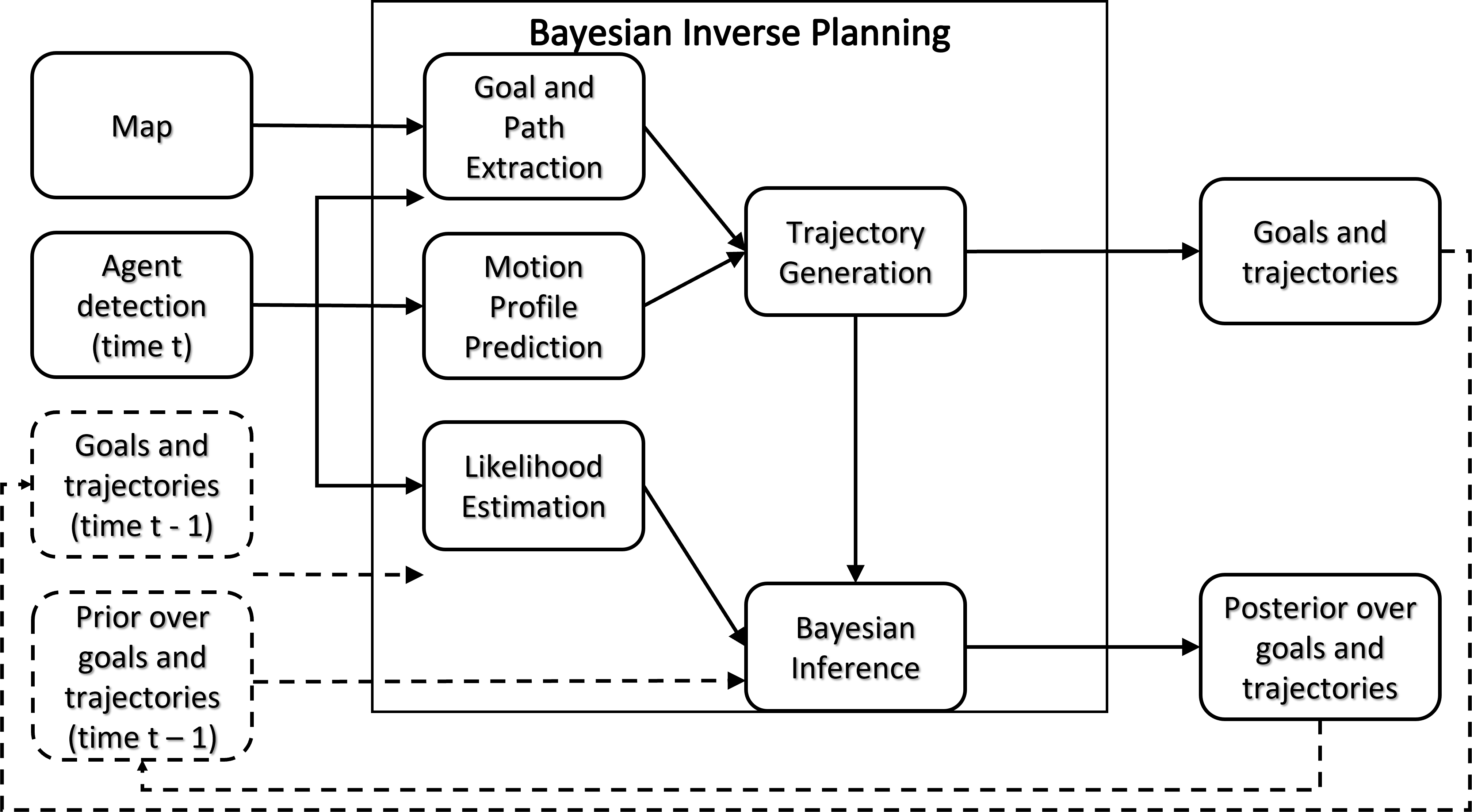}
 \caption{Bayesian Inverse Planning with Kinematic Models (BIP-KM) block diagram}
 \label{fig:BayesianInversePlanningOverview}
 \vspace{-0.5cm}
\end{figure}

An overview of the method is shown in \figurename~\ref{fig:BayesianInversePlanningOverview}. The model involves five components: 1) goal and path extraction, 2) motion profile prediction, 3) trajectory generation, 4) likelihood estimation and 5) Bayesian inference. Given a detected pose and a topological map that describes the geometry of roads and lanes, goals are extracted by exploring all lane graph traversals. For example, in highway situations, goals will correspond to staying in the lane, changing to a neighboring lane, entering the highway or exiting it. For each goal, reference paths can be extracted from the lane mid-lines and combined with target motion profiles to generate trajectories. In this implementation, a single goal-directed trajectory is generated for each goal/reference path, using the pure pursuit controller~\cite{coulter1992implementation}, which can enforce a bicycle model and kinematic limits, e.g. maximum steering and accelerations. The target acceleration profile relies on a decaying acceleration model since a constant acceleration model can be more accurate than a constant velocity model in the short term but less accurate in the long term~\cite{lefevre2014comparison}. The likelihood of each goal and trajectory is estimated by comparing each trajectory with the observed agent state. The Bayes rule is finally used to update the beliefs over the set of goals and trajectories by multiplying the likelihood with the prior to give the posterior, the model's prediction, while accounting for goal changes over time.

\subsection{Behavioral analysis}

\begin{figure}
 \vspace{-0.0cm}
 \centering
 \subfigure[]{
 \includegraphics[width = 0.45\linewidth, ]{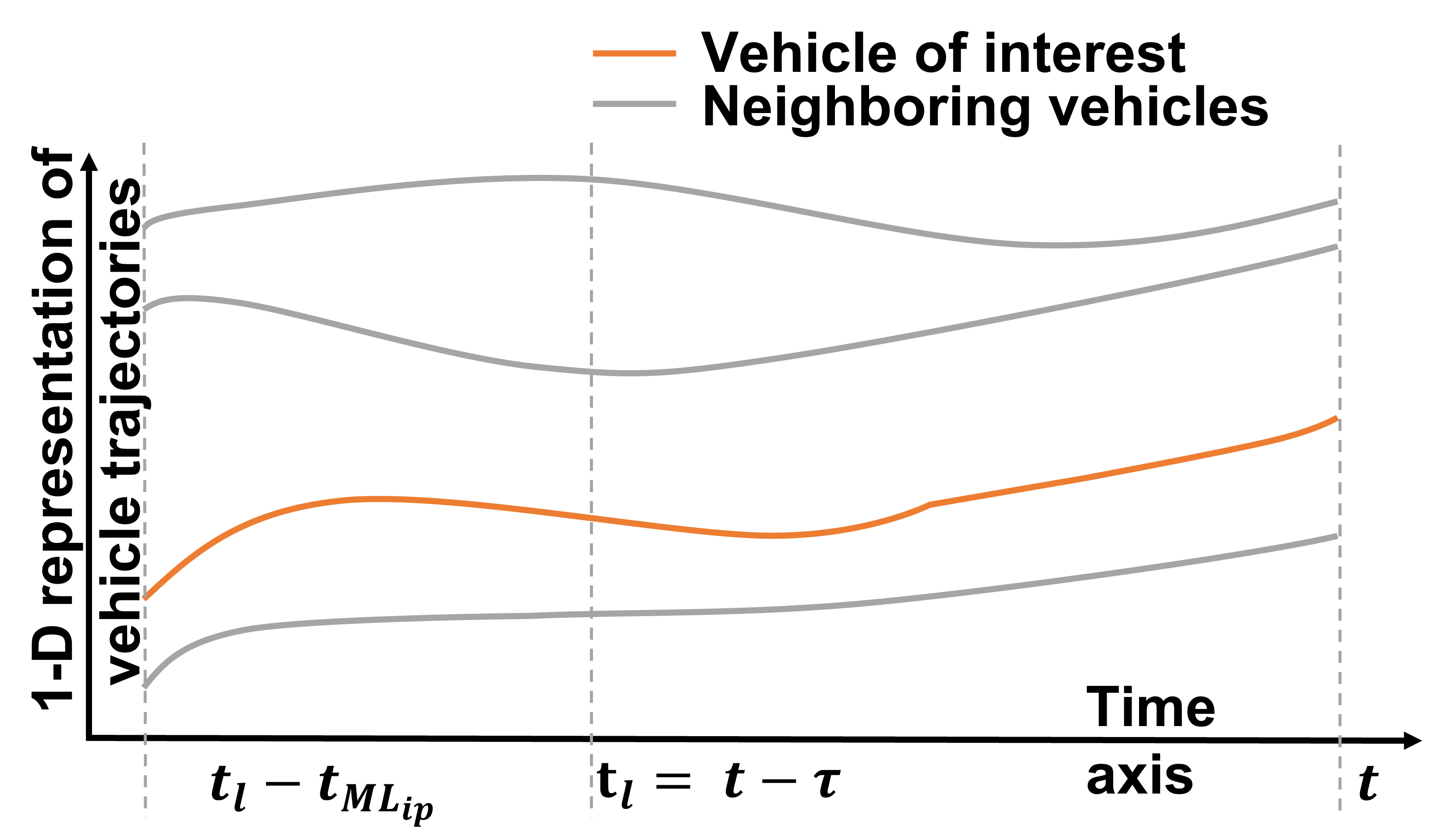} \label{fig:beh_analysis_sche_nat}}
 \subfigure[]{
 \includegraphics[width = 0.45\linewidth]{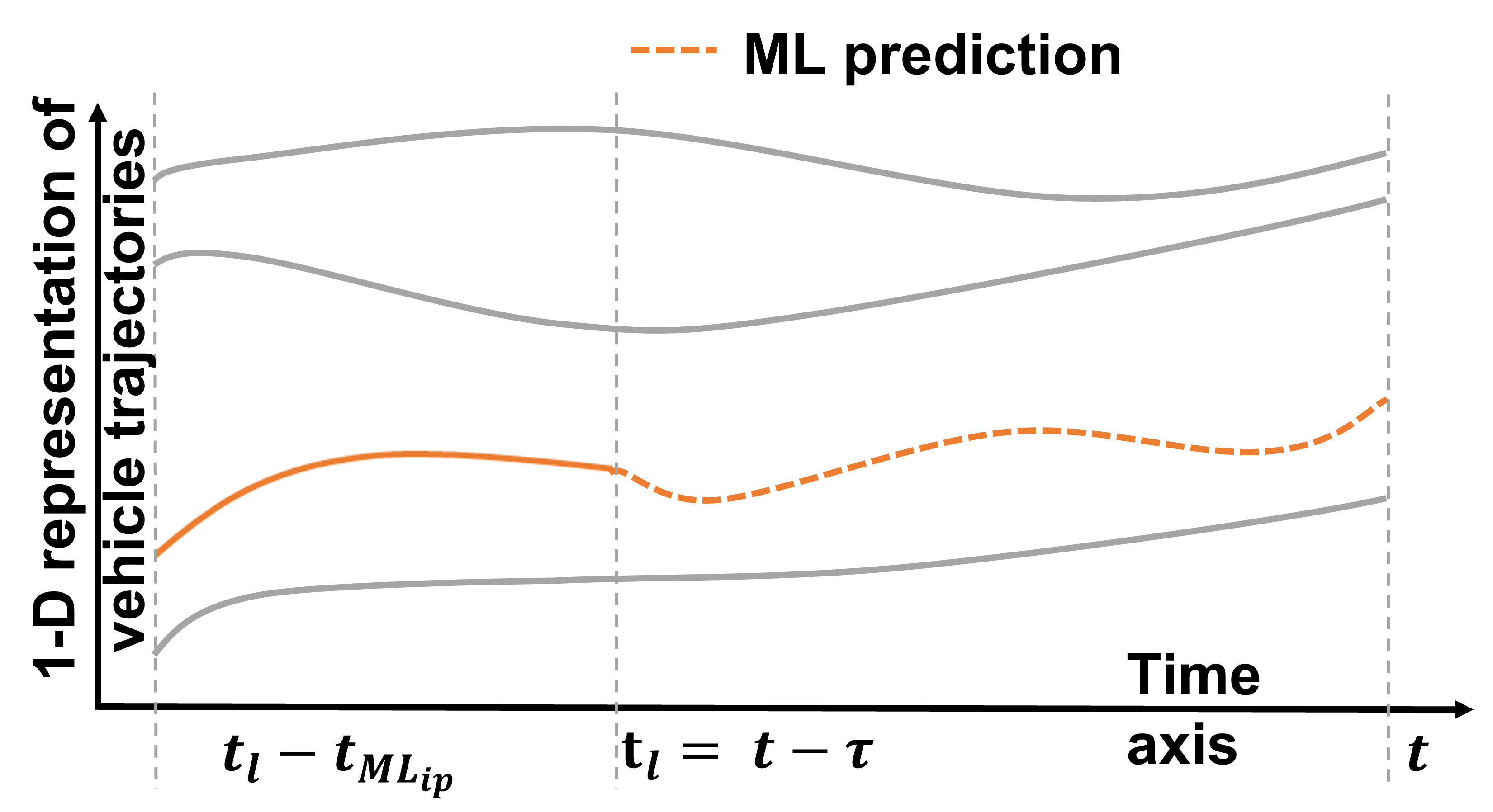} \label{fig:beh_analysis_sche_ml}}
 \caption{Schematics for behavior comparison between (a) naturalistic and (b) machine-learned trajectories prediction. Note: the trajectories for neighboring vehicles are same in both }
 \label{fig:beh-sche}
 \vspace{-0.6cm}
\end{figure}

\figurename~\ref{fig:beh-sche} shows a graphical representation of the behavioral comparison. We have used a one-dimensional trajectory to depict the schematic for ease of visual understanding. In reality, both the longitudinal and lateral positions of the vehicles were used for training the machine-learned models and in the behavioral analysis. All the machine-learned models were given a trajectory input of $3$ seconds from the appropriate time stamp for the vehicle of interest. The vehicle of interest is all the highway vehicles in the outermost lane of the highway and the vehicles in the merging lane of the highway in the case of merging scenario and all other vehicles on the highway (individually) in the case of highway lane change scenario. All the machine-learned models outputted $5$ seconds trajectory predictions for the vehicle of interest. The most likely predicted trajectory was used for the behavioral analysis. We divide the highway driving into two scenarios: merging lane driving and rest of the highway driving since they involve quite different interaction dynamics.

Humans judge collision conflicts by primarily relying on first-order motion information~\cite{markkula2016farewell} and can gauge the time-to-arrival (TTA) of objects advancing towards them~\cite{lee1976theory}. On the basis of these findings, we hypothesized that, with first-order kinematics, it is possible to observe many primary behavioral patterns during vehicle-vehicle interactions.

\subsubsection{Merging scenarios}
First, the vehicles involved in the merging scenario and the corresponding merging time, $t_m$, are extracted from the trajectory dataset. The merging time, $t_m$, is the time when the merging vehicle completed its transition into the highway lane from the on-ramp. We examine the kinematic history of the vehicle on the on-ramp and the vehicle on the outermost highway lane with a look-back window, $\tau$. The look-back window, $\tau$, helps us understand why the merge happened the way it happened, i.e. to see kinematic snapshots of the vehicles involved in the interaction at different time points in the past and understand why the interaction played out the way it did. All three machine-learned models involved in this study are capable of producing $5$ seconds of future trajectory given $3$ seconds of past trajectory as input (both ego-vehicle and neighboring vehicles). Since the machine-learned models predictions were for $5$ seconds, the look-back window was fixed from $1$ second up to $5$ seconds at $1$ second intervals from the merging time, $t_m$, for all the interacting pairs of vehicles.

\begin{figure}
 \centering
 \includegraphics[width=\linewidth]{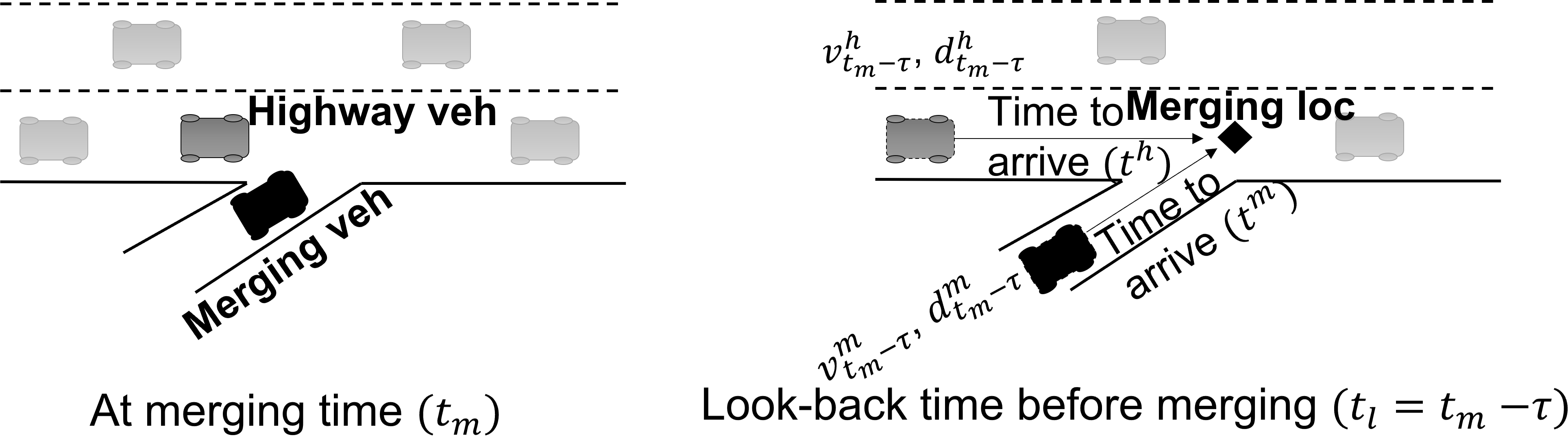}
 \caption{A graphical example of Bias for kinematically leading agent to pass the merging point first}
 \label{fig:KLA_illu}
\end{figure}

\begin{figure}
\centering
 \vspace{-0.3cm}
 \includegraphics[width = \linewidth]{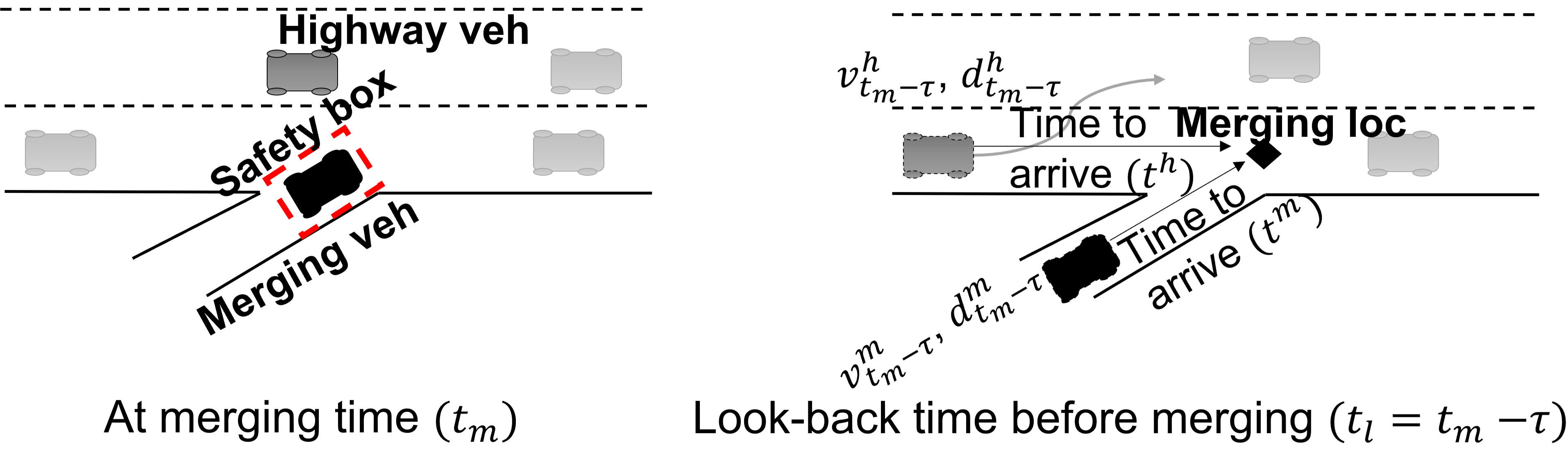}
 \vspace{-0.4cm}
 \caption{A pictorial representation of ``courtesy'' lane change by highway vehicle in merging scenario along with an illustration of safety box for collision aversion analysis}
 \label{fig:MLC_illu}
 \vspace{-0.6cm}
\end{figure}

For both the vehicle on-ramp and the highway vehicle, the interacting pair, we find their distance to merging point at a given look-back time, $t_l = t_m - \tau$, and compute the instantaneous velocities of the vehicle from the naturalistic driving data. Let $v^m_t, d^m_t$ and $v^h_t, d^h_t$ represent the instantaneous velocity and distance to the merging point for the on-ramp vehicle (merging vehicle, $x^m$) and the highway vehicle ($x^h$) respectively at any given time, $t$. We can then compute the lead time for the merging vehicle with respect to time to arrival of the highway vehicle at the merging point for a given look-back time with the formula $T_\tau = \frac{d^h_{t_l}}{v^h_{t_l}} - \frac{d^m_{t_l}}{v^m_{t_l}}$. This lead time for the merging vehicle was used to analyze the merging behaviors in the naturalistic data and the machine-learned models with the exact same procedure.
\paragraph{Bias for kinematically leading agent to pass first} The bias for kinematically leading agent to pass the merging point first is illustrated in \figurename~\ref{fig:KLA_illu}. From our definition, when the lead time, $T_\tau$ is positive it indicates the merging vehicle has a kinematic lead over the highway vehicle at that particular time and vice versa in case of negative lead time. In a scenario with two vehicles interacting, we expect that whichever agent has a clear lead, when $T_{\tau}$ is sufficiently large, the agent with the lead would always pass the merging point first~\cite{choudhury_modeling_2009,kang2017game}. Also, when the lead is not sufficient, more precisely as $T_\tau$ approaches zero, this pattern should break down.

\paragraph{Courtesy lane change to yield} \figurename~\ref{fig:MLC_illu} depicts a hypothetical lane change by the highway vehicle. In the illustration, it is assumed that the highway vehicle changed lane in the look-back time, $t_l = t_m-\tau$, until the merging time, $t_m$, to facilitate the on-ramp vehicle merging into the highway to make our analysis depiction clearer. In our analysis, we counted all lane changes that happened in that window in the naturalistic and ML predicted trajectories since these lane changes can be considered as a potential courtesy lane change to accommodate the merging/on-ramp vehicle(s). We expect the frequency of lane changes to increase when there is space-sharing conflict. In other words, if the lead time for the merging vehicle, $T_\tau$ is close to zero we expect the probability of lane change by the highway vehicle to increase in order to accommodate the merge and also to avoid collision.

\begin{figure}
 \centering
 \includegraphics[width=\linewidth]{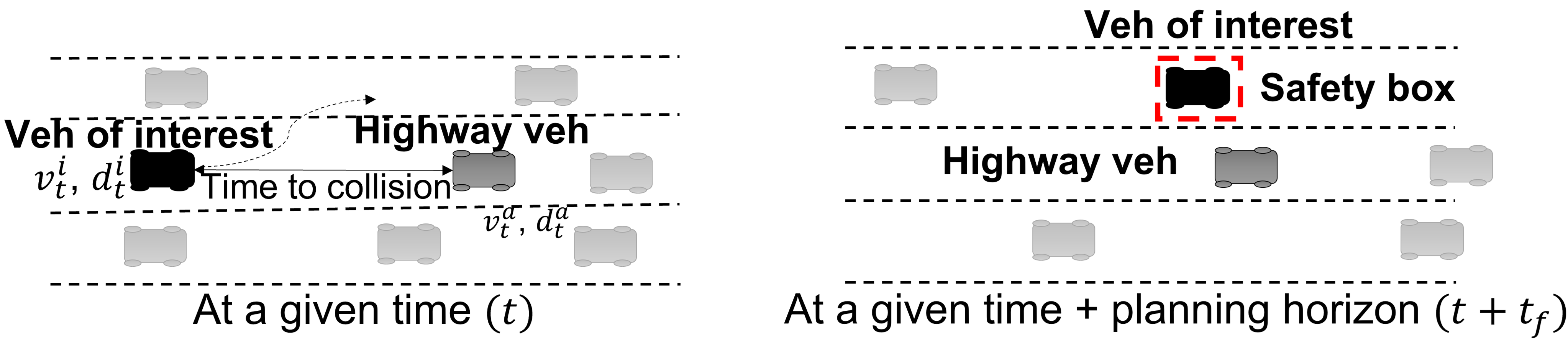}
 \caption{An illustration of highway lane changes into nominal faster lane}
 \label{fig:HLC_illu}
\end{figure}

\begin{figure}
 \centering
 \includegraphics[scale =0.20]{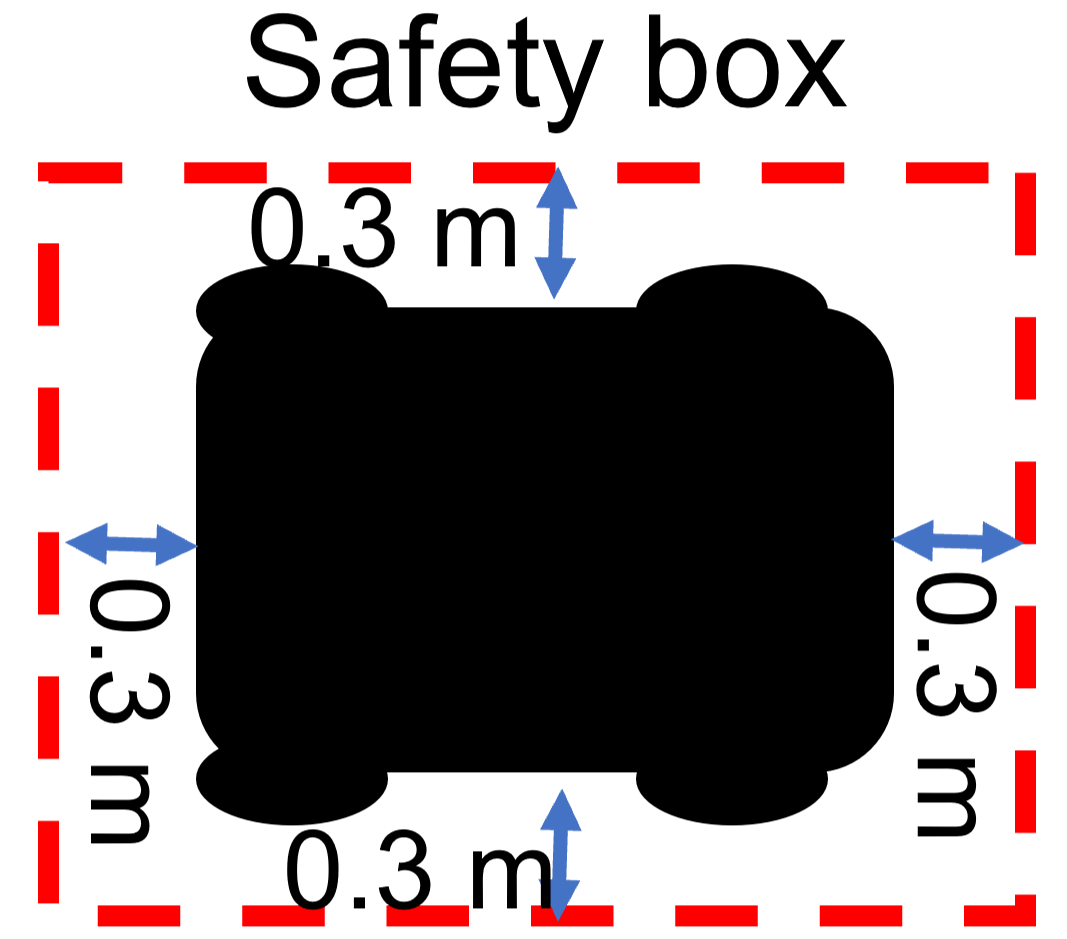}
 \caption{A depiction of safety box for collision aversion analysis. A vehicle is augmented with $0.3$ meters of additional space in all direction and it is considered an unsafe interaction if any other vehicle encroaches into this space.}
 \label{fig:safety_illu}
\vspace{-0.5cm}
\end{figure}

\subsubsection{Highway driving scenario - lane changes within highway}
For vehicles driving on highways, we are interested in lane change scenarios where the highway vehicle is switching into the nominal faster lane. We consider the primary interacting agent with a given vehicle at any time point to be the one in front of it to keep the behavioral analysis tractable. Thus, the instantaneous time to collision, $T_{ttc}$, between ego-vehicle ($x^e$) and the vehicle immediately in front ($x^f$) is of interest. The instantaneous positions and velocities ($d^e_t, d^f_t, v^e_t, v^f_t$) of both vehicles can be obtained from the naturalistic data. Thus, the instantaneous time to collision $T_{ttc}$ can be computed with the formula $T_{ttc} = (d^f_t-d^e_t)/(v^e_t-v^f_t)$. The time to collision for each vehicle was used to analyze the highway lane change behavior in both the naturalistic data and the machine-learned models with the exact same procedure.

The most likely cause for within-highway lane changes can be the desire to avoid imminent collision or deceleration~\cite{jula2000,7728060}. \figurename~\ref{fig:HLC_illu} shows an imaginary highway lane change. We use the time to collision $T_{ttc}$ measure between the vehicle of interest and the vehicle immediately ahead of it to analyze the lane change behavior within highway driving. It is expected that when the time to collision $T_{ttc}$ approaches zero the frequency of lane changes will increase.

\subsubsection{Collision aversion behavior}
Finally, we are interested in the collision avoidance behavior of predicted trajectories in comparison to the naturalistic data which the machine-learned models are trained on. We extended the bounding box of individual vehicle dimension with an additional safety margin of $0.3$ meters ($1$ feet) in all directions for all vehicles at all time instances as illustrated in \figurename~\ref{fig:safety_illu}. We obtain the safety statistics by counting the number of times a bounding box for a given vehicle is intersected/overlapped by another vehicle's bounding box. In the naturalistic trajectory statistics, all the vehicles' instantaneous positions are from the recorded naturalistic data. In the machine-learned models' statistics, the ego vehicle's instantaneous position is from machine-learned models' prediction and rest of the vehicles' instantaneous position is from the naturalistic driving data. All vehicles at all timestamp are considered individually as ego vehicle for the counting statistics. We expect the machine-learned models to perform at different levels since the models rely on different motion models, especially CSP-LSTM and BIP-KM do not enforce physical safety while IPBT assesses collisions to account for interactions explicitly. Overall, we expect the human drivers to only very rarely exhibit this type of near-collision behavior. In other words, we expect the human drivers to be collision averse.
\section{RESULTS \& DISCUSSION}
\subsection{RMSE}

\begin{table}
 \caption{Trajectory error on the NGSIM test set}
 \centering
 \begin{tabular}{c c c c c}
 \hline
 \specialcell{Evaluation \\Metric} & \specialcell{Prediction \\Horizon (s)} & \specialcell{CSP-LSTM} & \specialcell{IPBT} & \specialcell{BIP-KM}\\ [0.2ex]
 \hline
 \multirow{5}*{RMSE (m)}& 1 & 0.5693 & 1.1579 & 0.5868 \\
 & 2 & 1.2575 & 1.9652 & 1.5540 \\
 & 3 & 2.1010 & 2.9808 & 2.8134 \\
 & 4 & 3.1674 & 4.2165 & 4.3747 \\
 & 5 & 4.4872 & 5.6221 & 6.2149 \\
 \hline
 \label{tab:RMSETable}
 \end{tabular}
 \vspace{-0.6cm}
\end{table}

The root mean square error (RMSE) at different time points in the predicted trajectories is presented in \tablename~\ref{tab:RMSETable}. All the evaluations, both RMSE and behavioral analysis, are based on the most likely trajectory, ignoring other trajectories. Using average trajectory error metrics like RMSE is a traditional way to compare machine-learned models, and it is used to justify a model's capability or improvement over existing models. The RMSE for the CSP-LSTM model is comparable to the one reported in the original paper~\cite{deo2018convolutional}. CSP-LSTM obtains lower RMSEs with differences when compared with IPBT and BIP-KM ranging from 1.75 cm to 1.73 m. Differences can vary across time steps. For example, BIP-KM has lower RMSEs than IPBT for shorter prediction horizons (1, 2, 3 seconds) and IPBT has lower RMSEs than BIP-KM for longer prediction horizons (4 and 5 seconds). Differences are often in the order of centimeters; for example, when comparing CSP-LSTM with BIP-KM at 1 and 2 seconds predictions, the differences are 1.75 cm and 29.65 cm respectively, or when comparing IPBT with BIP-KM at 3 and 4 seconds predictions, the differences are 16.74 cm and 15.82 cm. Aggregate metrics like RMSE can hide important patterns, especially in unbalanced dataset containing rare events, and the resulting differences can be very difficult to interpret. As we will see in the rest of this section, an analysis rooted on behavioral metrics gives a more in-depth insight into the model capabilities. The high-level metric does not capture nor describe behavioral differences among the machine-learned models.

\subsection{Bias for kinematically leading agent to pass first in merging scenarios}\label{sec:kla_result}

\begin{figure}
 \centering
 \subfigure[]{
 \includegraphics[width = 0.45\linewidth]{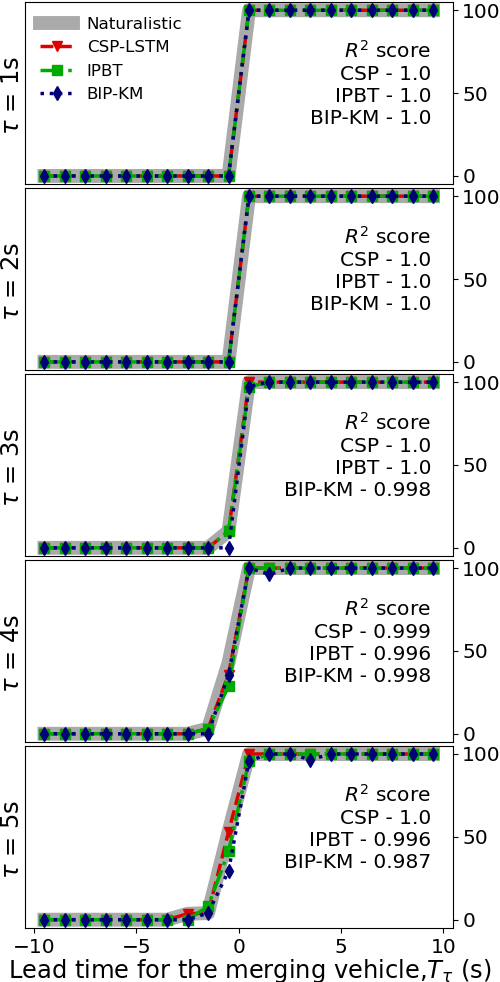} \label{fig:kla_US101}}
 \subfigure[]{
 \includegraphics[ width = 0.45\linewidth]{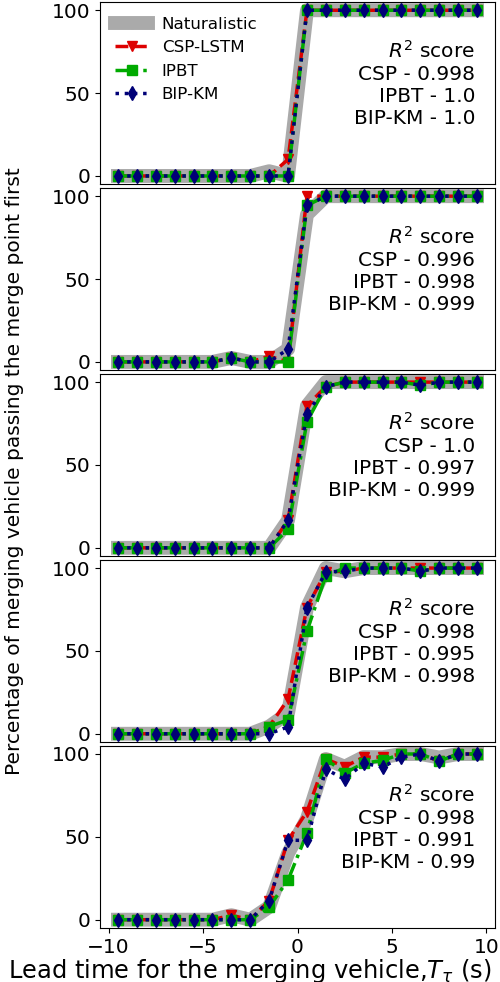} \label{fig:kla_I80}}
 \caption{Merging lane - Bias for kinematically leading agent to pass first in (a) US101 highway (b) I80 highway. All the models have captured this behavior similar to the pattern exhibited in naturalistic driving.}
 \label{fig:KLA}
 \vspace{-0.6cm}
\end{figure}

In \figurename~\ref{fig:KLA}, a positive lead time for the merging vehicle for a certain look-back time ($\tau$) before the merge indicates that the merging vehicle had kinematic edge over the highway vehicle at that time instance, while negative lead time for the merging vehicle indicates the opposite. In the naturalistic data, we can observe a pattern of kinematically leading agents always passing the merging point first. This is in line with our expected human behavior. It is also evident from the tightly overlapping curves in \figurename~\ref{fig:KLA} that all the models are able to capture this first-order kinematic dependent behavior. We computed coefficient of determination, $R^2$, to check how well the machine-learned models trajectory predictions captured the behavior observed in the naturalistic driving. A $R^2$ value of $1$ indicates that the model has captured the behavior exactly as in the naturalistic driving, and $0$ or negative values indicate complete failure. All the models, in both the highways (US101 and I80), have a $R^2$ value greater than $0.99$. This indicates that the bias for kinematically leading agent to pass first behavior has been captured accurately from the naturalistic driving by all 3 models. Even though average trajectory error metrics can lead to the conclusion that some models are better than others, from the bias for kinematically leading agent to pass first behavioral perspective, all the 3 models are performing equally, including BIP-KM which predicts each agent independently. This is in contrast to comparison based on just average quantitative metric, which does not allow us to determine whether the models are capturing this behavioral phenomenon.

\subsection{Courtesy lane change to yield in merging scenarios}\label{sec:mlc_result}

\begin{figure}
 \centering
 \subfigure[]{
 \includegraphics[width = 0.45\linewidth]{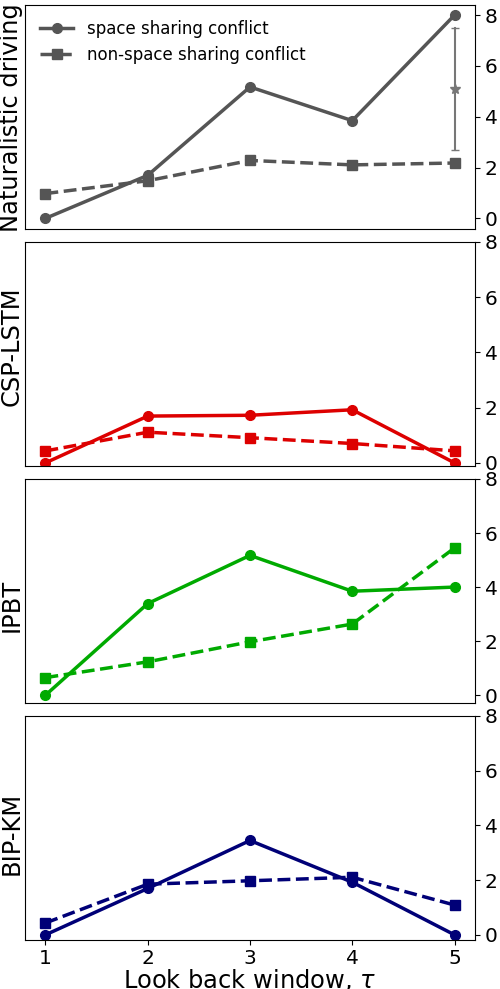} \label{fig:mlc_US101}}
 \subfigure[]{
 \includegraphics[width = 0.45\linewidth]{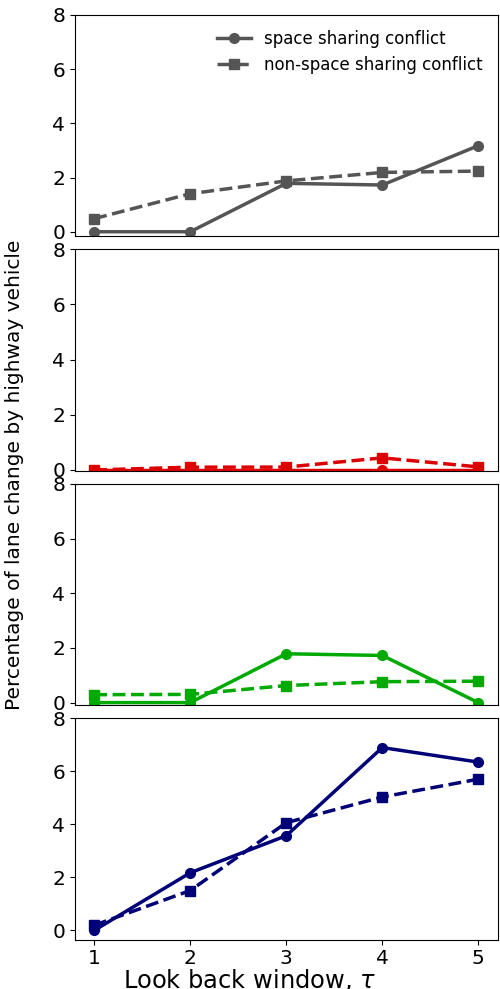} \label{fig:mlc_I80}}
 \caption{Merging lane - ``courtesy'' lane changing behavior statistics. Comparison between the trends in space-sharing conflict situation and non space-sharing conflict situation in (a) US101 Highway and (b) I80 Highway. The vertical bar in the US101 $\tau = 5 s$ indicates statistically significant difference (p-value $0.0367$) between the lane change in space-sharing conflict situation and other situations based on Fisher's exact test. The models struggle at varying degree to capture this more nuanced behavior.}
 \label{fig:MLC}
 \vspace{-0.6cm}
\end{figure}

In the ``courtesy'' lane change to yield in merging situation behavior, we are interested in the lane changes that happened when there was potential space-sharing conflict and otherwise (no conflict situations) (\figurename~\ref{fig:MLC}). When the vehicle merging on to the highway and the highway vehicle on the outermost lane are within $\pm 1$ second with respect to their arrival at the merging location we deem it as a space-sharing conflict situation. We did a Fisher's exact test to determine whether there was significant difference in the frequency of lane change behavior between situations with and without space-sharing conflicts. In other words, when there is an emerging space-sharing conflict, we expect the probability of lane change by highway vehicle to increase. We found that the lane change behavior in US101 highway by human drivers with a look-back window of 5 seconds was significantly different (p-value $0.0367$) between the space-sharing conflict situations and other situations, \figurename~\ref{fig:mlc_US101}. This is not captured by any of the three machine learned models. They did not return a statistical significance between the space-sharing conflict lane changes and non space-sharing conflict situation lane changes. It is worth noting that predicting longer horizons is difficult for all predictors. Even though the look-back window ($\tau$) was $5$ seconds it does not mean that the lane change happened a long time into the predicted future, it just means that there was a long-range (in distance) interaction between the two vehicles. The highway vehicle performs a lane change when there is still almost $5$ seconds left until the merge conflict, and none of the models seem capable of learning this long-range interaction. I80 does not have this statistical significance at any time stamp in the naturalistic driving since it was a more congested highway leading to less opportunity for a lane change.

\begin{figure}
 \centering
 \subfigure[]{
 \includegraphics[ scale =0.18]{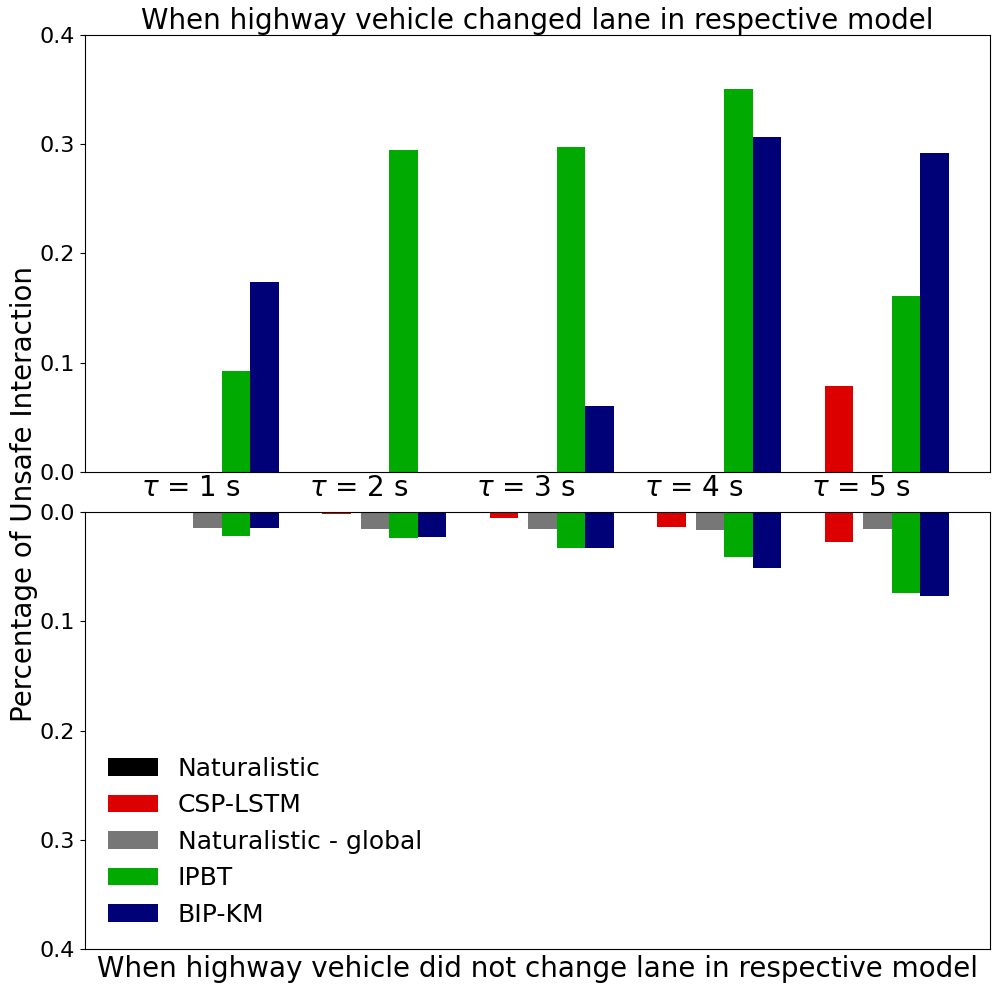} \label{fig:mlc_US101_safety}}
 \subfigure[]{
 \includegraphics[ scale =0.18]{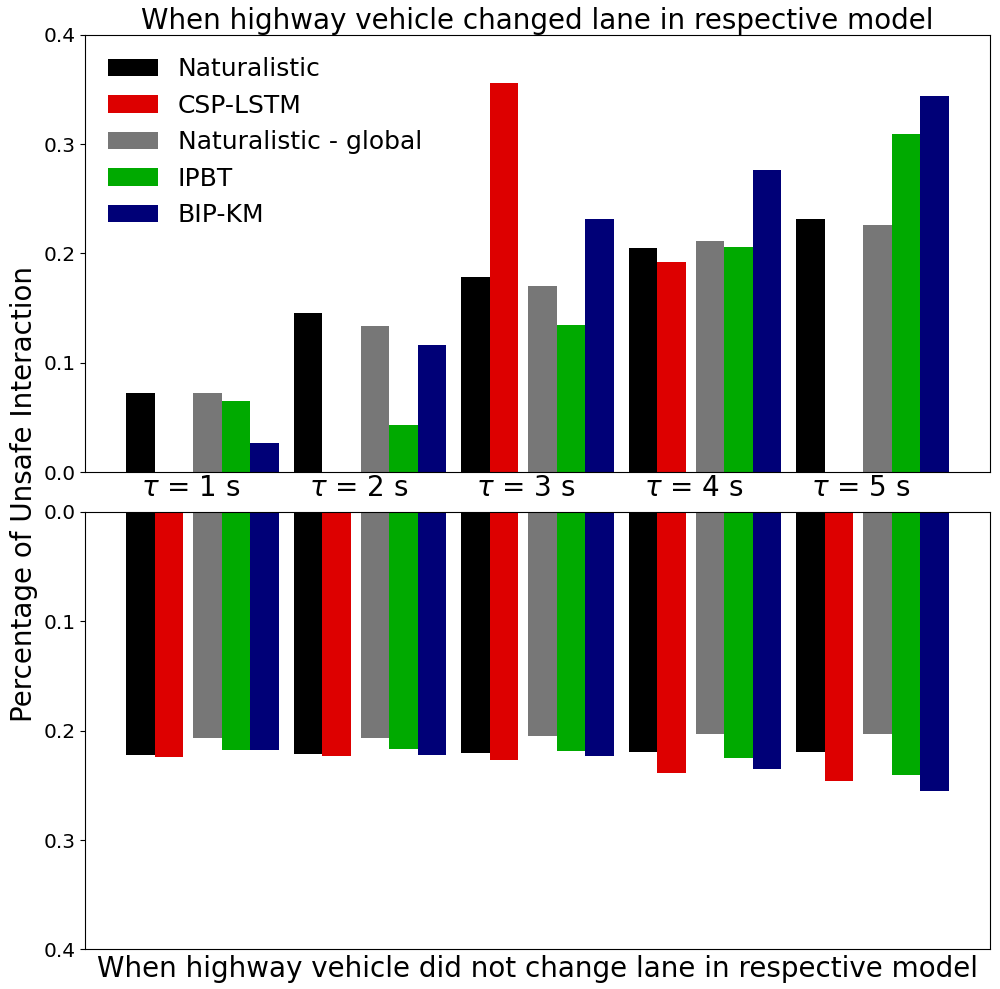}
 \label{fig:mlc_I80_safety}}
 \caption{Merging lane - collision aversion summary statistics within lane changing behavior based on bounding box fitted onto each vehicle in (a) US101 highway and (b) I80 highway. The y-axis is the percentage of total interaction that were deemed unsafe (bounding box overlapping) when vehicle did change or did not change lane in the respective models. Since y values are below 1.0\,\%, the y-axis is truncated for clarity.}
 \label{fig:MLC_time_safety}
 \vspace{-0.6cm}
\end{figure}

\begin{figure*}
 \centering
 \subfigure[]{
 \includegraphics[scale = 0.28] {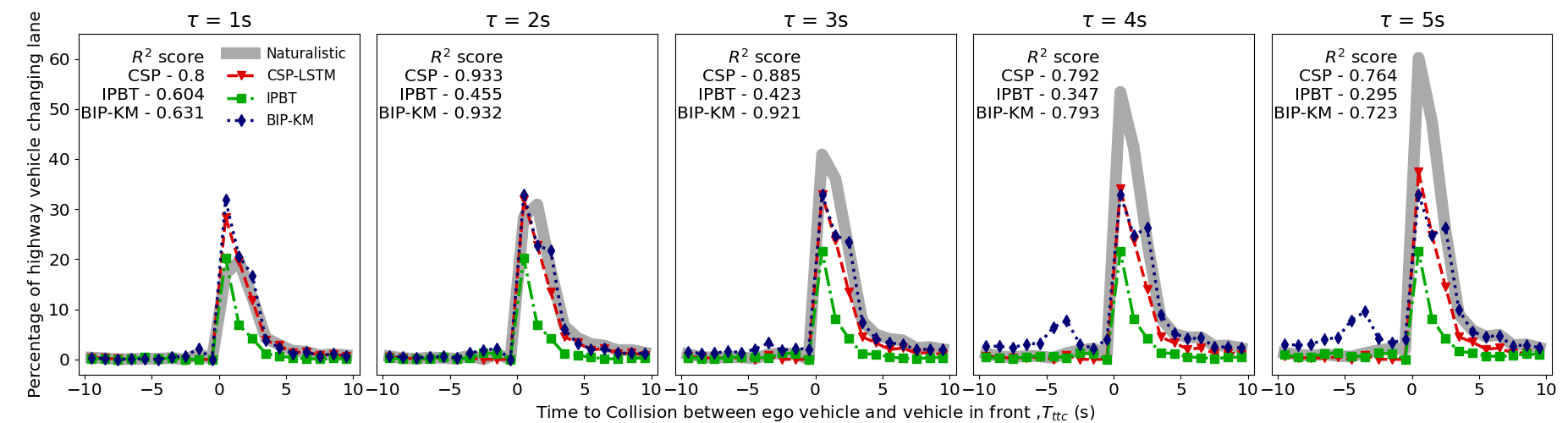} \label{fig:hlc_US101}
 \vspace{-0.25cm}
 }
 \subfigure[]{
 \includegraphics[scale = 0.28]{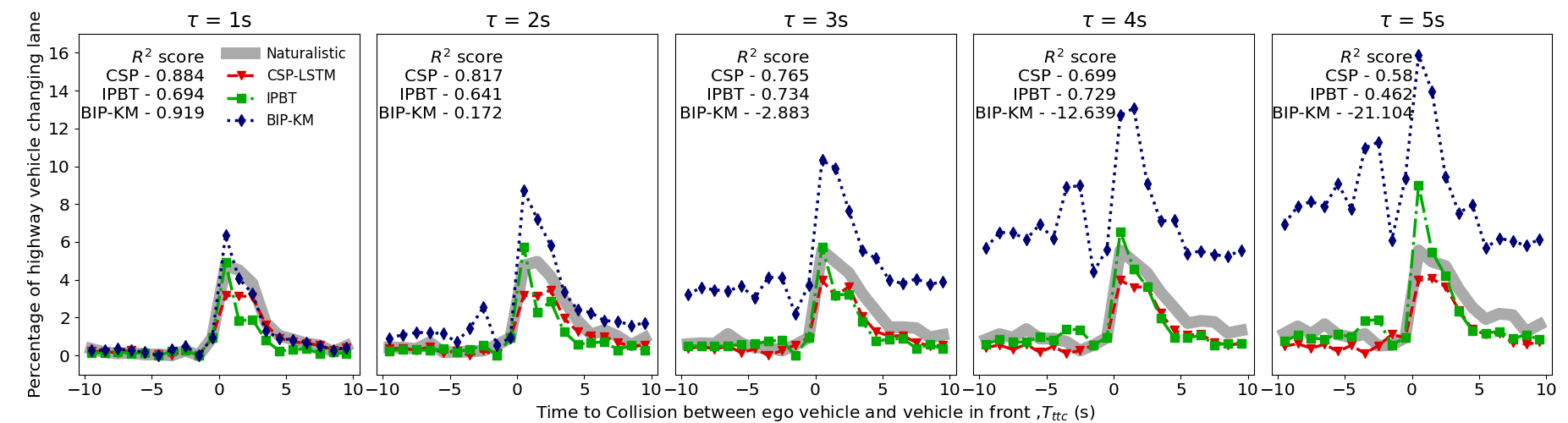}
 \label{fig:hlc_I80}
 \vspace{-0.6cm}
 }
 \caption{Highway Driving - lane changing into the nominal faster lane behavior statistics in (a) US101 highway and (b) I80 highway. The models again struggle to capture the lane change behavior in highway driving similar to the merging scenario ``courtesy'' lane changes.}
 \label{fig:HLC}
 \vspace{-0.6cm}
\end{figure*}

Additionally, we can see in \figurename~\ref{fig:MLC} that the human lane changes generally increased with increasing look-back window in both the highways, but none of the models is able to fully capture this pattern. As shown in \figurename~\ref{fig:MLC}, the models have different lane change prediction capabilities, not faithfully captured when using high-level average error metric like RMSE. In both highways, CSP-LSTM captures very few lane changes, always less than IPBT and BIP-KM. The behavior is captured or missed at various degrees by the models. For example, IPBT seems to capture it very well in I80 for shorter horizon ($\tau <= 4$ s) in space-sharing conflict situations. In general, for all methods, longer horizons look difficult. The high-level average error quantitative metric does not reflect the lane change behavior in detail like this. To humans, courtesy lane changing is a salient and important behavior pattern. The measure of differences between the predicted trajectories and the actual trajectories is not sufficient to determine whether this lane changes behavior to accommodate merging vehicles has been captured explicitly. This discrepancy in capturing the ``courtesy'' lane changes clearly shows the need for models which take into account human behavioral aspect.

Next, we checked the safety of the predicted trajectories with respect to all the other vehicles' recorded naturalistic driving trajectories (\figurename~\ref{fig:MLC_time_safety}). In particular, we compared the collision aversion behavior of humans versus the machine-learning models predictions. Since the CSP-LSTM predictions were based on the local coordinates and the other two models, IPBT and BIP-KM predictions were based on the global coordinates from the NGSIM dataset~\cite{NGSiMData}, we have reported the collision aversion behavior with respect to both coordinate systems\footnote{There will be slight difference between the local coordinates and global coordinates in the naturalistic driving. This is due to the fact that the NGSIM dataset provider's algorithmic coordinate conversion is not exactly a one-to-one mapping~\cite{NGSiMData}.}. Collision or safety statistics is generally presented as secondary results and not discussed in detail in machine-learning literature~\cite{gu2021densetnt,ngiam2021scene}. Here, we instead discussed this metric directly and as a follow-up to our behavioral analysis. We are interested in the safety performance of the models both when they recommended lane change and when they did not compared to the naturalistic driving.

The number of unsafe interactions in US101 highway (\figurename~\ref{fig:mlc_US101_safety}) is of lesser magnitude than the I80 highway (\figurename~\ref{fig:mlc_I80_safety}) in naturalistic driving. This could be attributed to the fact that I80 was the more congested highway and thus there were more chances for vehicles to be within each other's designated safety space. In \figurename~\ref{fig:MLC_time_safety}, neither the lane change predictions of the models (top half of the graphs) nor the no lane change predictions (bottom half of the graphs) of the models could match the naturalistic data in terms of total number of detected unsafe interaction. CSP-LSTM performed poorly in the more congested highway, I80 (\figurename~\ref{fig:mlc_I80_safety}). In US101 highway, the BIP-KM model recommended generally less lane changes than IPBT model (\figurename~\ref{fig:mlc_US101}) and in I80 highway, the BIP-KM model recommended generally more lane changes than IPBT model (\figurename~\ref{fig:mlc_I80}). Nevertheless, the collision aversion behavior of BIP-KM was equivalent to IPBT model in most cases (\figurename~\ref{fig:mlc_US101_safety},~\ref{fig:mlc_I80_safety}). This shows that BIP-KM has captured implicitly to some extent the collision aversion behavior in merging scenarios. Overall, the collision aversion analysis within merging scenarios reinforces the conclusion that the models have struggled to capture the physical and interactive aspect of ``courtesy'' lane change behavior.

\subsection{Highway lane change behavior}\label{sec:hlc_result}
The highway lane change behavior is a behavior of interest from a behavioral modeler's perspective. When the time to collision between an ego vehicle and the vehicle in front is positive, it means the ego vehicle is catching up with the vehicle in front and, if no action is taken by either vehicle, it can lead to collision in the near future. In this situation we expect the probability of lane change by the ego vehicle to increase. On the other hand, a negative time to collision indicates the vehicle in front faster than the ego vehicle at that instance. In this scenario, we expect less lane change by the ego vehicle since there is no collision risk with the vehicle in front. In \figurename~\ref{fig:HLC}, we can see that the highway lane change by vehicles into the nominal faster lanes (from outermost to innermost lane in the direction of travel) is following the expected pattern in the naturalistic driving. In US101 highway, the less congested highway, with increasing time window (left to right in the \figurename~\ref{fig:hlc_US101}) and relatively low time to collision we can see an uptick in lane changes by the highway vehicles. In I80 highway (\figurename~\ref{fig:hlc_I80}), the more congested highway, a steady amount of lane changes is exhibited in the naturalistic driving regardless of the time window. This lane change behavior, in I80 highway, could be due to the physical feasibility of lane changes being relatively low in a congested highway.

\figurename~\ref{fig:HLC} shows the highway vehicle lane change behavior from the naturalistic driving reproduced in the models prediction at different extent in both the highways. This is evident from the $R^2$ values. In US101 highway (\figurename~\ref{fig:hlc_US101}), CSP-LSTM and BIP-KM reproduce the behavior considerably for all time windows. But, IPBT depicts a steady lane change characteristics across all time windows in US101 highway similar to the naturalistic driving highway lane change behavior in I80 highway. In I80 highway (\figurename~\ref{fig:hlc_I80}), CSP-LSTM and IPBT are the better performing model across different time windows compared to the BIP-KM predictions. The plots show that, in congested scenarios, BIP-KM tends to assign higher probabilities to lane changes than the other models. One possible explanation for failure of all the models in I80 highway except time window $1$ second could be that in a congested highway, the interactions involved before and after a lane change to maintain kinematic advantage or to avoid safety related incidents are more complex compared to a non-congested highway.

\begin{figure}
 \centering
 \subfigure[]{
 \includegraphics[scale =0.18]{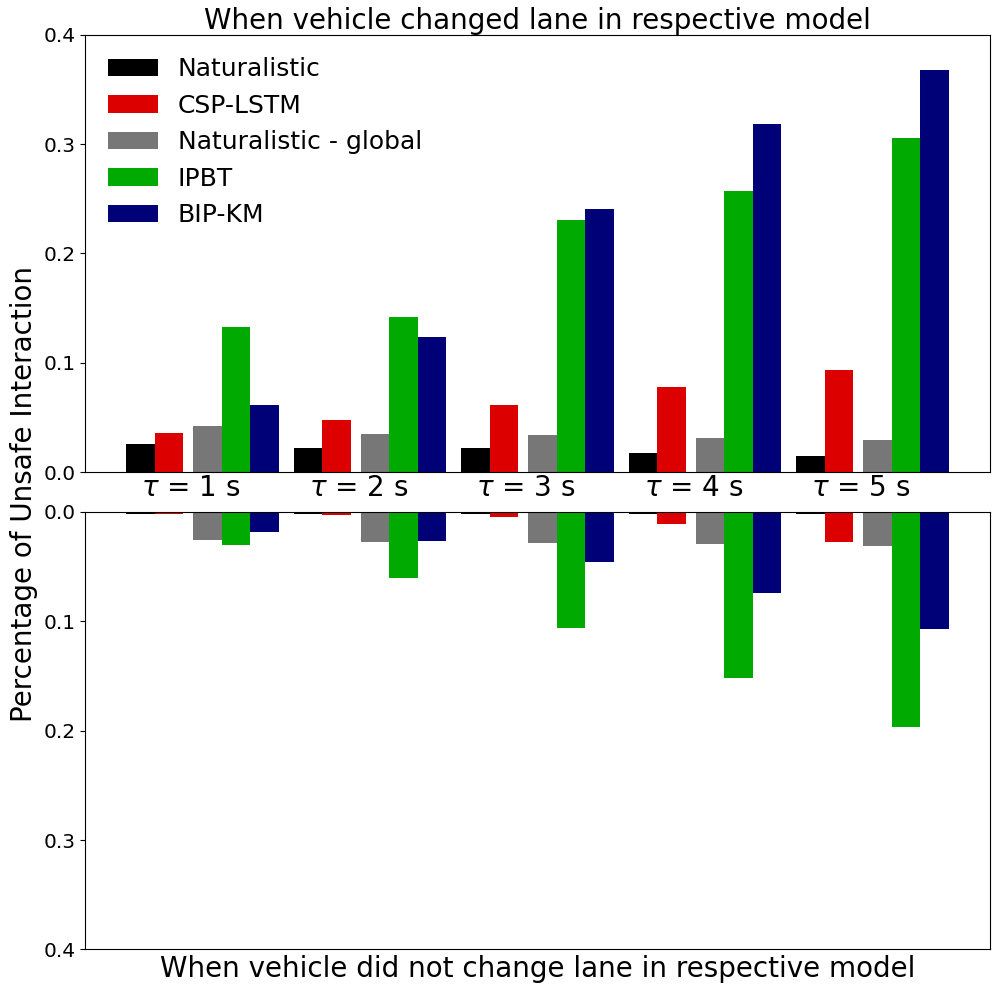} \label{fig:hlc_US101_safety}}
 \subfigure[]{
 \includegraphics[ scale =0.18]{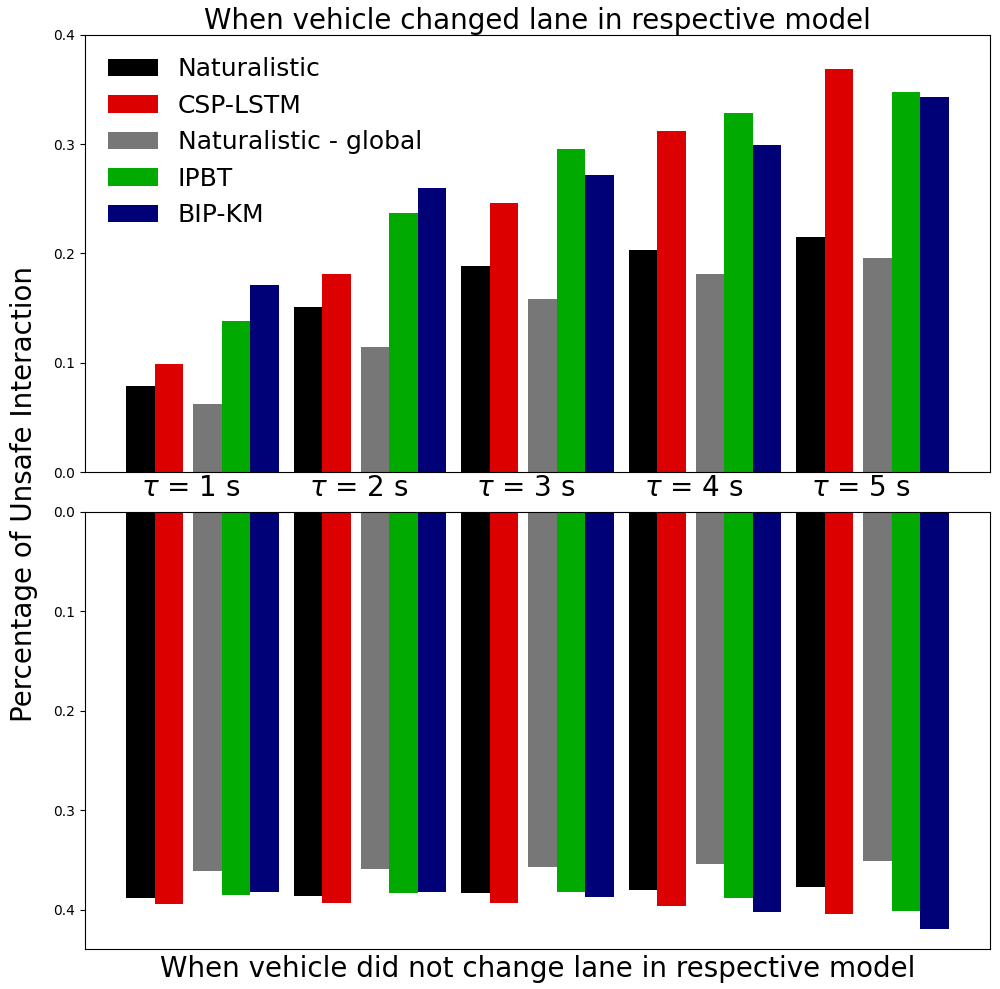}
 \label{fig:hlc_I80_safety}}
 \caption{Highway driving - collision aversion statistics within lane changing behavior based on bounding box fitted onto each vehicle in (a) US101 highway and (b) I80 Highway. The y-axis is the percentage of total interaction that were deemed unsafe (bounding box overlapping) when vehicle did change or did not change lane in the respective models. Since y values are below 1.0\,\%, the y-axis is truncated for clarity.}
 \label{fig:HLC_safety}
 \vspace*{-0.6cm}
\end{figure}

The machine-learned models either captured the behavior as a constant independent of the time window or repeated the pattern learned in one highway to another which is not true in the ground truth data. Overall, all the models struggled at varying degrees, and our behavioral analysis helps showing behavioral differences that cannot be deduced directly with high-level average trajectory error metrics (RMSE, mFDE, mADE).

We performed collision aversion analysis for all the vehicles driving on highways excluding the ones already accounted for in the merging scenario. Recall that unsafe interaction is defined as coming within $0.3$ meters of another vehicle in any direction. Also, an unsafe interaction can happen due to a vehicle getting within the safety margins of the vehicle in front (due to lane change/no lane change by the vehicle), or getting within the safety margins of the vehicle in adjacent lanes (due to self or other vehicle's lane change or no lane change). Similar to merging scenarios, there is discrepancy between US101 and I80 highway's collision aversion behavior in naturalistic trajectories (\figurename~\ref{fig:HLC_safety}). This again can be attributed to the congested nature of I80 compared to US101 highway. Comparing safety statistics between the naturalistic driving and the machine-learned models' predictions, the models' prediction generally perform worse across all three models. Overall none of the models outperform or match the human collision aversion behavior in the highway driving. The insight into models' highway lane change behavior prediction and subsequent collision aversion characteristic cannot be obtained by just a high-level metric comparison. In general, our analysis and results point towards the usefulness of behavior-oriented holistic approach to evaluation and comparison of machine-learned trajectory predictors.

\section{CONCLUSION AND FUTURE WORKS}
The aim of this work was to investigate the extent to which machine-learned models capture human-like traffic behavior in a highway driving scenario. We started with highway driving as our behavioral benchmark, hypothesized a number of different human interaction phenomena that could be present in naturalistic highway driving and developed mathematical analysis methods that allowed us to confirm the presence of the expected human behavioral patterns in the naturalistic driving dataset (NGSIM). We then applied the exact same analysis methods to the trajectories predicted by three different machine-learned models trained with that dataset. All three models were trained with the same subset of the data, and the behavioral analysis of human data and model prediction was done using the validation and the test subset of the dataset. Recapping our results and discussion in brief: Section~\ref{sec:kla_result} showed that the three machine-learned trajectory prediction methods are comparable despite making different assumptions/approach to solve the problem. Section~\ref{sec:mlc_result} and~\ref{sec:hlc_result} showed that models have different lane change prediction capabilities and they struggle capturing the interactive aspect of the lane change behavior. Also, they can exhibit different behaviors in the two highways with different congestion levels. This indicates that the tested machine-learned models are able to capture some behaviors, like kinematically leading agent passing the merging point first, but struggle to capture behaviors that are arguably more nuanced, like courtesy lane changes and lane changes to maintain a kinematic edge. These model shortcomings could be either due to their formulation or lack of explicit focus on capturing these behavioral phenomena, as evidenced in~\cite{siebinga2021validating}.

Our results demonstrate that it is hard to tell from just a high-level quantitative error metric like RMSE in what ways a model's behavior is human-like or not. Analyzing the output of our behavioral metrics can aid model development too. The main takeaway from our findings is not in the specific results obtained for this dataset and these models, but rather that these findings allow us to conclude that the question asked in the Introduction, "How low an average trajectory error is low enough?" is fundamentally ill-posed. We need behavioral analysis of the trajectory predictions to report about the behavioral competence of a given model. Thus, we argue for a richer, behavior-anchored quantitative analysis on top of traditional quantitative metrics, to understand a model's behavioral capabilities from a human perspective. Our work also opens new interesting questions about how these behavior-anchored quantitative metric(s) can be integrated into machine-learning of behavior models, for example as part of the loss function, as additional regularization parameters, or by guiding the design of the model itself by better knowledge of human behavior.

Another promising future research direction is to include different kinds of traffic actors and driving environments to test the behavioral competence of the machine-learned models in a wider range of conditions. As we had mentioned in our problem formulation, we restricted ourselves to highway driving to keep our problem definition feasible. We have now shown that even in the restricted behavioral space, the tested models do not perform adequately from a behavioral perspective.

\ifCLASSOPTIONcaptionsoff
 \newpage
\fi

\bibliographystyle{IEEEtran}
\bibliography{srinivasan_beyond_rmse.bib}%

\begin{IEEEbiography}[{\includegraphics[width=1in,height=1.25in,clip]{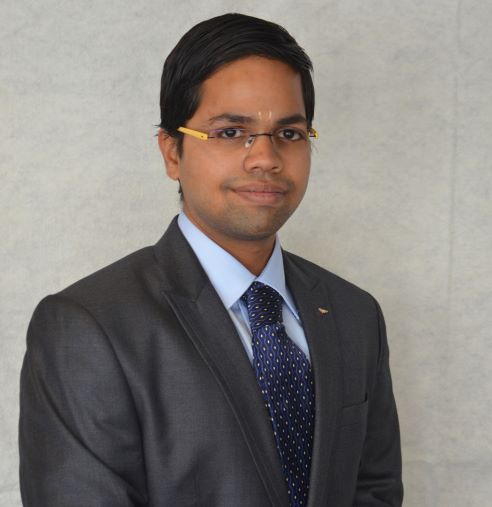}}]{Aravinda Ramakrishnan Srinivasan}
received the B.Tech. degree in electronics and communication engineering from the SASTRA University, Tirumalaisamudram, Tamil Nadu, India, and the M.S. and Ph.D. degrees in mechatronics and mechanical engineering from the University of Tennessee, Knoxville, TN, USA. Before joining the Human Factors and Safety group at Institute for Transport Studies, University of Leeds, UK as research fellow, he was a postdoctoral research fellow at the Lincoln Centre for Autonomous Systems, University of Lincoln, UK. His research interests include machine-learning, artificial intelligence, autonomous vehicles, and robotics applications in everyday life.
\end{IEEEbiography}%

\begin{IEEEbiography}[{\includegraphics[width=1in,height=1.25in,clip]{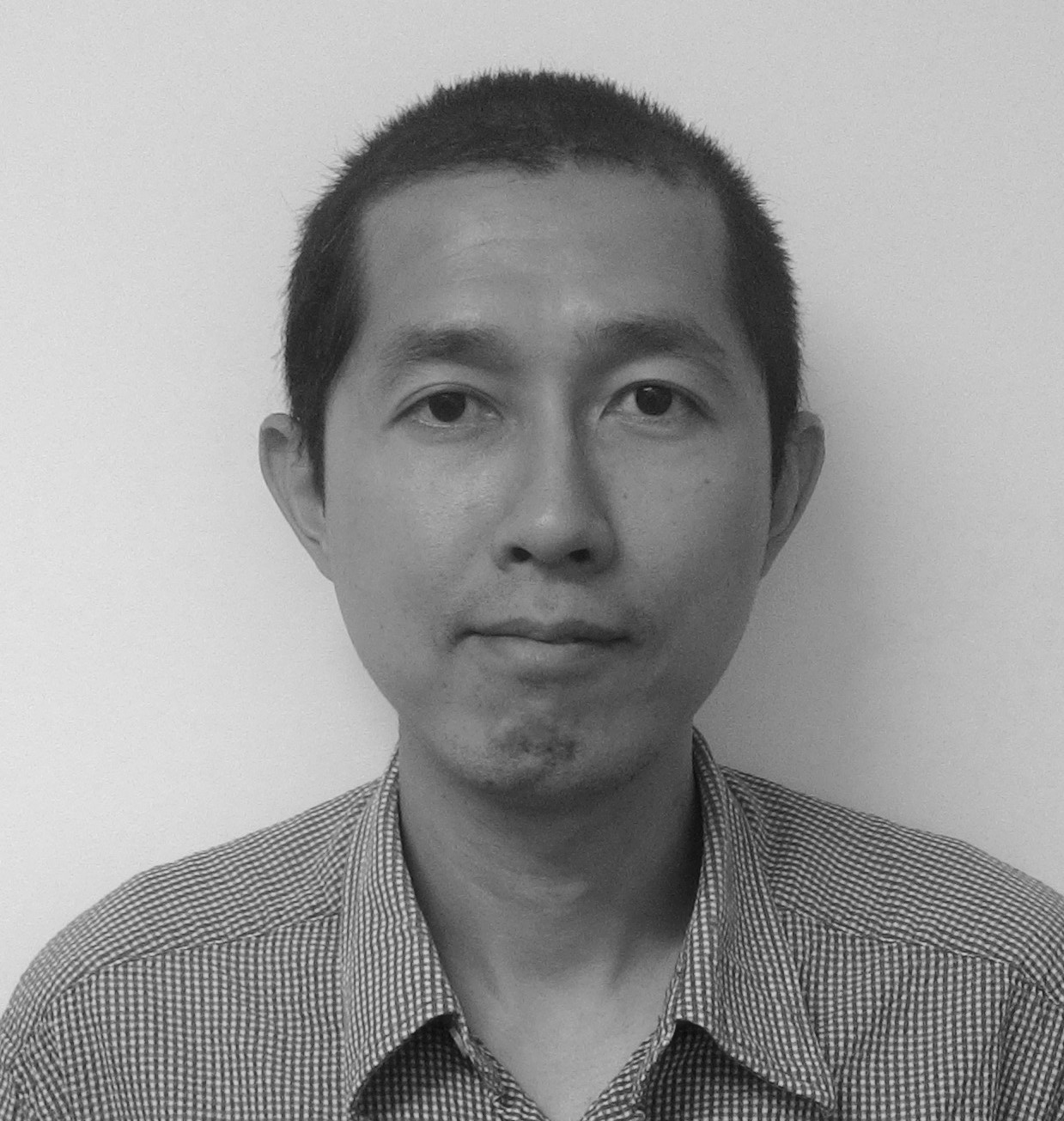}}]{Yi-Shin Lin}. received the B.S. degree in psychology from National Taiwan University, Taipei, Taiwan, in 2002, the M.A. degree in psychology from the City University of New York, New York City, U.S.A., in 2007 and the Ph.D. degree in experimental psychology from University of Birmingham, Birmingham, U.K. in 2015. From 2015 to 2018, he was a post-doctoral researcher with Tasmania Cognition Lab, University of Tasmania, Australia. He joins the Institute for Transport Studies, the University of Leeds, in 2020. His research interests cover three-facets: Human cognition, traffic psychology, and research methods. These include decision theory, high-performance computing, simulation-based numerical methods, computational modeling, Bayesian inference, and road safety.
\end{IEEEbiography}%

\begin{IEEEbiography}[{\includegraphics[width=1in,height=1.25in,clip]{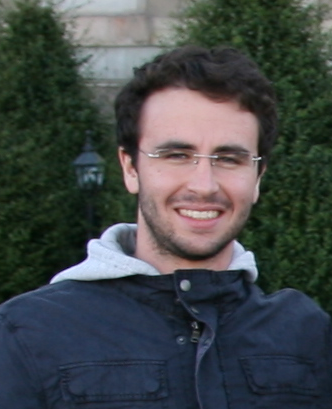}}]{Morris Antonello} is a Lead Research Engineer at Five AI, Edinburgh. He received a PhD in Information Science and Engineering from the University of Padova in 2018. He was a visiting researcher at the Technische Universität Wien in 2016. He is interested in motion prediction and planning, computer vision and their applications in autonomous vehicles and robotics.
\vspace*{1\baselineskip} 
\end{IEEEbiography}%

\begin{IEEEbiography}[{\includegraphics[width=1in,height=1.25in,clip]{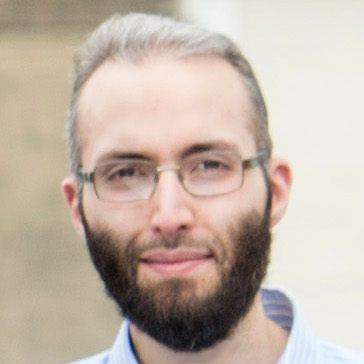}}]{Majd Hawasly} is a Scientist at the Qatar Computing Research Institute at Hamad Bin Khalifa University, Qatar. Before that, he was a Lead Research Scientist in the Motion Planning and Prediction Applied Research team at Five. He received his PhD from the School of Informatics at the University of Edinburgh in 2014. After that, he was a postdoctoral research fellow at the University of Leeds.
\end{IEEEbiography}%

\begin{IEEEbiography}[{\includegraphics[width=1in,height=1.25in,clip]{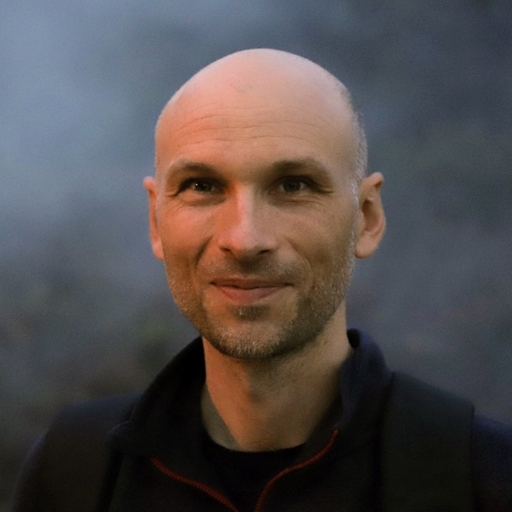}}]{Anthony Knittel} completed a PhD in cognitive science and computer vision from the University of New South Wales, and has researched game-playing topics at the Center for the Mind, and computer vision for human motion recognition at Canon Information Systems Research Australia. He is currently working on motion prediction related topics for autonomous vehicles at Five and Bosch. He is interested in autonomous systems and how they can be informed by human cognitive processes.
\end{IEEEbiography}%

\begin{IEEEbiography}[{\includegraphics[width=1in,height=1.25in,clip]{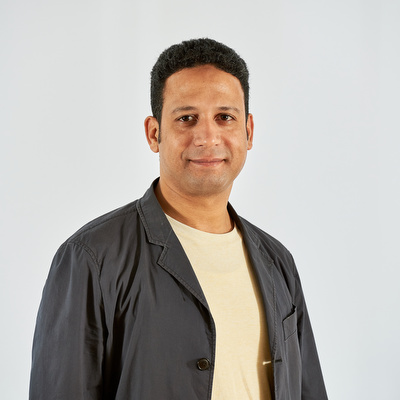}}]{Mohamed Hasan} is a Machine Learning Scientist at Gaist Solutions, UK. Before Gaist, he was a Research Fellow at the University of Leeds. He received a PhD in Robotics Engineering from Egypt-Japan University of Science and Technology and was a post-doc at Osaka University. His research interests include robot manipulation planning, visual localization and mapping, and motion planning of autonomous vehicles.
\end{IEEEbiography}%

\begin{IEEEbiography}[{\includegraphics[width=1in,height=1.25in,clip]{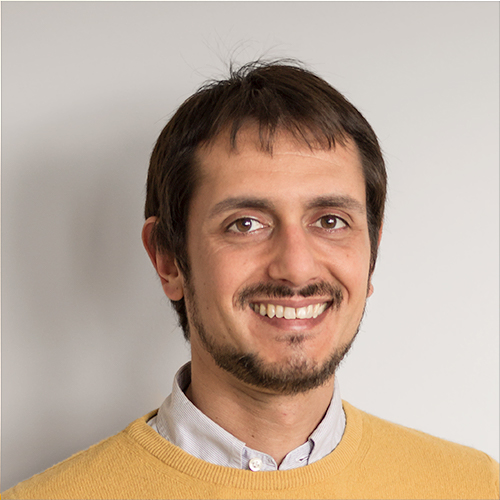}}]{Matteo Leonetti} is a Lecturer in Autonomous Systems at King's College London. Before King's, he was a lecturer at the University of Leeds. He received a PhD in Computer Engineering from Sapienza University of Rome and was a post-doc at the Italian Institute of Technology and the University of Texas at Austin. His research interests include reinforcement learning, planning and reasoning, and autonomous robots.
\end{IEEEbiography}%

\begin{IEEEbiography}[{\includegraphics[width=1in,height=1.25in,clip]{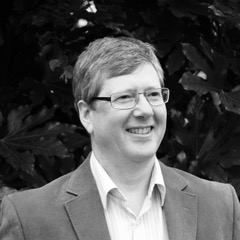}}]{John Redford} is CTO and a co-founder of Five AI, a start-up recently acquired by Bosch, which is dedicated to developing safe autonomous driving systems. He has more than 35 years of software development experience spanning autonomous vehicles, machine learning, signal processing, processor architecture and operating systems. Previously he has held positions including Distinguished Engineer and Director Engineering at Broadcom, Co-founder and VP Software of Element 14, and CTO of Acorn Computers. He holds an MA in Mathematics from the University of Cambridge.
\end{IEEEbiography}%

\begin{IEEEbiography}[{\includegraphics[width=1in,height=1.25in,clip]{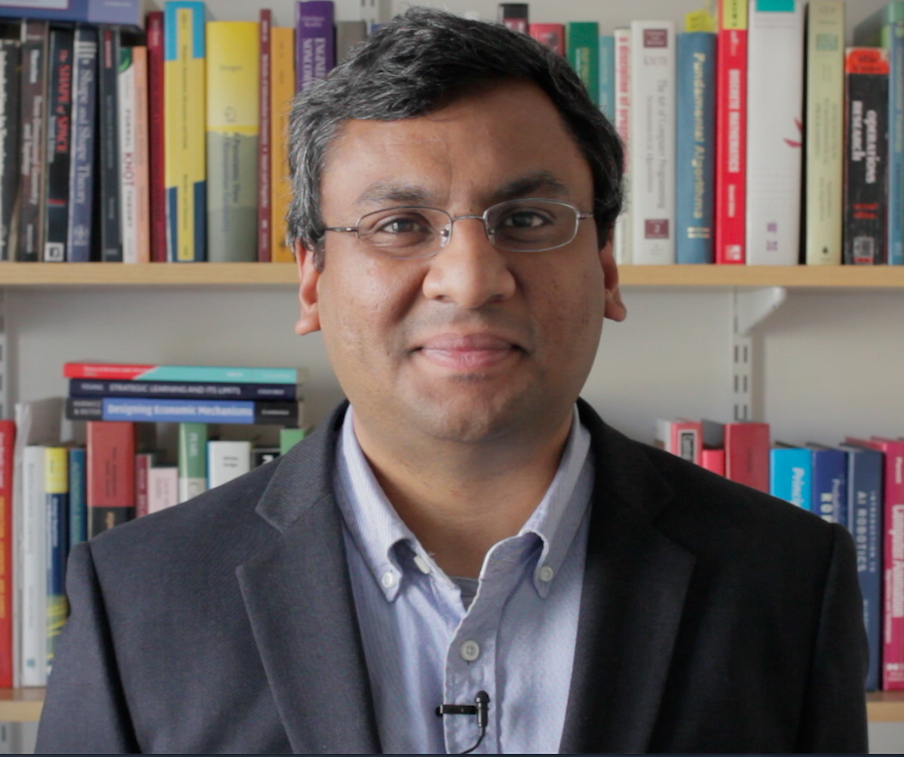}}]{Subramanian Ramamoorthy} is a Professor of Robot Learning and Autonomy in the School of Informatics at the University of Edinburgh, where he is also Director of the Institute of Perception, Action and Behaviour, and founding Member of the Executive Committee for the Edinburgh Centre for Robotics. He is a Turing Fellow at the Alan Turing Institute and Member of the UK Computing Research Committee convened by BCS and IET. He received his PhD in Electrical and Computer Engineering from The University of Texas at Austin in 2007. He has been a Member of the Young Academy of Scotland at the Royal Society of Edinburgh, and has held Visiting Professor positions at the University of Rome “La Sapienza” and at Stanford University. His research investigates learning, adaptation, and control mechanisms that enable autonomous robots to cope with the uncertain and the unknown, such as when working in human-centred environments. Between 2017-2020, he served as Vice-President – Prediction and Planning – at Five AI, a UK based company developing a technology stack for autonomous vehicles. He continues to be involved with the company as a Scientific Advisor.
\end{IEEEbiography}%

\begin{IEEEbiography}[{\includegraphics[width=1in,height=1.25in,clip]{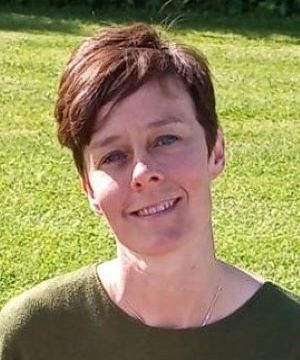}}]{Jac Billington} is an Associate Professor in Psychology at the University of Leeds. She completed a PhD in neuroscience at the University of Cambridge in 2007 and subsequently worked for six years as a postdoctoral research fellow at Royal Holloway, University of London. Her research focus is in understanding how people extract information from the surrounding environment for the purpose of voluntary action and successful self-motion. She is particularly interested in the neuroscientific underpinnings of such behaviours.
\end{IEEEbiography}%

\begin{IEEEbiography}[{\includegraphics[width=1in,height=1.25in,clip]{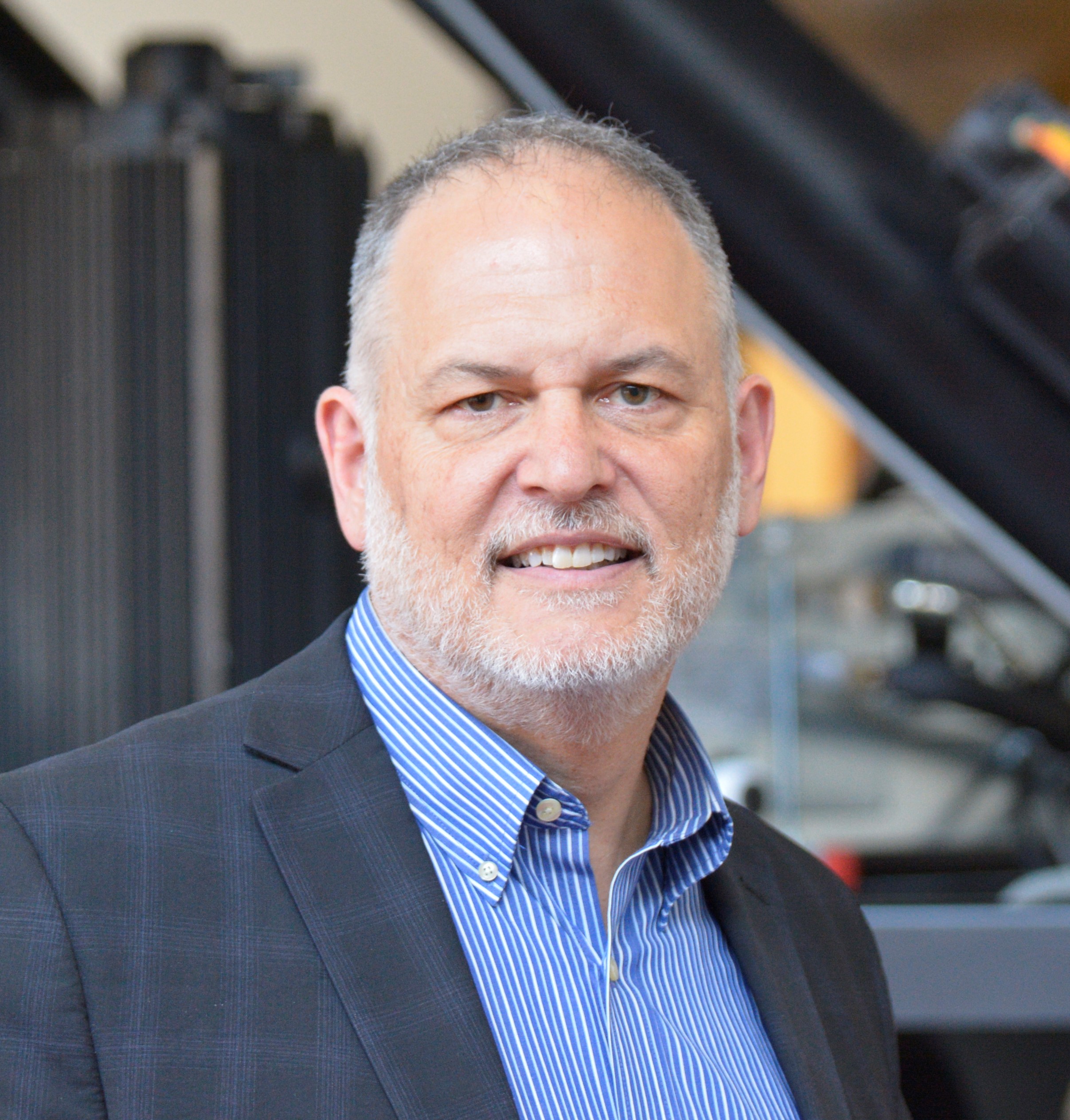}}]{Richard Romano} has over thirty years of experience developing and testing AVs and ADAS concepts and systems which began with the Automated Highway Systems (AHS) project while he directed the Iowa Driving Simulator in the early 1990’s. He received his BASc and MASc in Engineering Science and Aerospace Engineering respectively from the University of Toronto, Canada and a PhD in Motion Drive Algorithms for Large Excursion Motion Bases, Industrial Engineering from the University of Iowa, USA. In addition to a distinguished career in industry he has supervised numerous research projects and authored many journal papers. In 2015 he was appointed Leadership Chair in Driving Simulation at the Institute for Transport Studies, University of Leeds, UK. His research interests include the development, validation and application of transport simulation to support the human-centered design of vehicles and infrastructure.
\end{IEEEbiography}%

\begin{IEEEbiography}[{\includegraphics[width=1in,height=1.25in,clip]{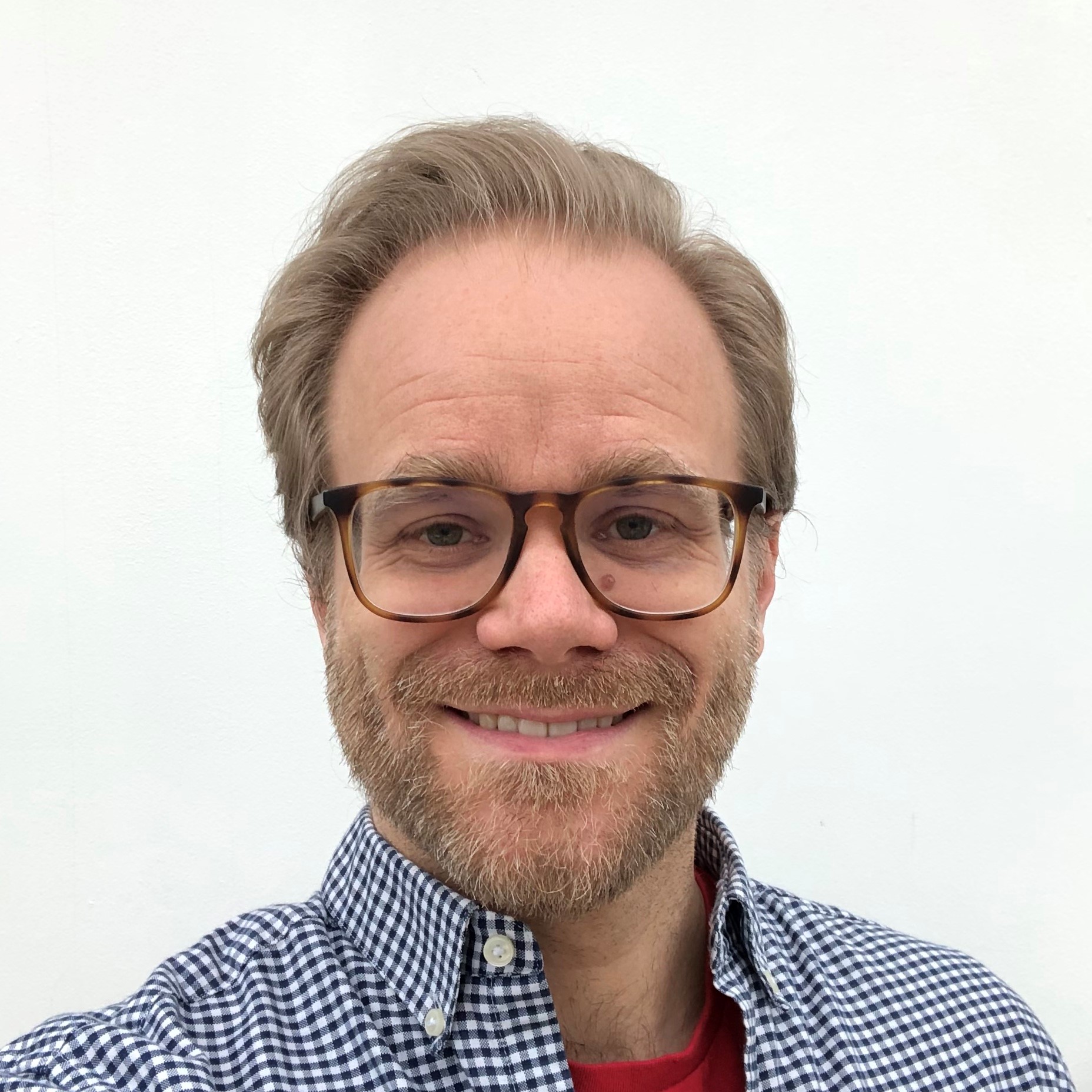}}]{Gustav Markkula} received the M.Sc. degree in engineering physics and complex adaptive systems and the Ph.D. degree in machine and vehicle systems from Chalmers University of Technology, Gothenburg, Sweden, in 2004 and 2015, respectively. He has more than a decade of research and development experience from the automotive industry, and is now Chair in Applied Behaviour Modelling at the Institute for Transport Studies, University of Leeds, UK. His current research interests include quantitative modeling of road user behavior and interaction, and virtual testing of vehicle safety and automation technology.
\end{IEEEbiography}%

\end{document}